\documentclass[10pt,twocolumn,letterpaper]{article}

\usepackage[pagenumbers]{cvpr} 
\usepackage{multirow}
\usepackage{bm}
\usepackage{makecell}
\usepackage{comment}
\usepackage{colortbl}
\usepackage[accsupp]{axessibility} 
\usepackage[dvipsnames]{xcolor}

\definecolor{firstcolor}{RGB}{199, 85, 93}
\definecolor{secondcolor}{RGB}{63, 217, 167}
\definecolor{thirdcolor}{RGB}{39, 87, 146}
\definecolor{firstcolor}{RGB}{252, 235, 193}
\definecolor{secondcolor}{RGB}{241, 233, 223} 

\newcommand{\markfirst}[1]{\colorbox{firstcolor}{\textbf{#1}}}
\newcommand{\marksecond}[1]{\colorbox{secondcolor}{#1}}

\definecolor{cvprblue}{rgb}{0.21,0.49,0.74}
\usepackage[pagebackref,breaklinks,colorlinks,citecolor=cvprblue]{hyperref}
\def\paperID{7944} 
\def\confName{CVPR}
\def\confYear{2024}

\title{360Loc: A Dataset and Benchmark for Omnidirectional Visual Localization with Cross-device Queries
}

\author{Huajian Huang$^{1*}$ \quad Changkun Liu$^{1*}$ \quad  Yipeng Zhu$^1$ \quad Hui Cheng$^2$ \quad  Tristan Braud$^1$ \quad  Sai-Kit Yeung$^1$\\
$^1$The Hong Kong University of Science and Technology \quad
$^2$Sun Yat-sen University \\
* equal contribution\\
{\tt\small \{hhuangbg, cliudg, yzhudg\}@connect.ust.hk},
{\tt\small chengh9@mail.sysu.edu.cn},
{\tt\small \{braudt, saikit\}@ust.hk}
}

\begin{document}
\maketitle
\begin{abstract}

Portable 360$^\circ$ cameras are becoming a cheap and efficient tool to establish large visual databases. By capturing omnidirectional views of a scene, these cameras could expedite building environment models that are essential for visual localization. However, such an advantage is often overlooked due to the lack of valuable datasets. This paper introduces a new benchmark dataset, 360Loc, composed of 360$^\circ$ images with ground truth poses for visual localization. We present a practical implementation of 360$^\circ$ mapping combining 360$^\circ$ images with lidar data to generate the ground truth 6DoF poses. 360Loc is the first dataset and benchmark that explores the challenge of cross-device visual positioning, involving 360$^\circ$ reference frames, and query frames from pinhole, ultra-wide FoV fisheye, and 360$^\circ$ cameras. We propose a virtual camera approach to generate lower-FoV query frames from 360$^\circ$ images, which ensures a fair comparison of performance among different query types in visual localization tasks. We also extend this virtual camera approach to feature matching-based and pose regression-based methods to alleviate the performance loss caused by the cross-device domain gap, and evaluate its effectiveness against state-of-the-art baselines. We demonstrate that omnidirectional visual localization is more robust in challenging large-scale scenes with symmetries and repetitive structures. These results provide new insights into 360-camera mapping and omnidirectional visual localization with cross-device queries. Project Page and dataset: \url{https://huajianup.github.io/research/360Loc/}.





\end{abstract}    
\section{Introduction}
\begin{figure*}
    \centering
    \includegraphics[width=\textwidth]{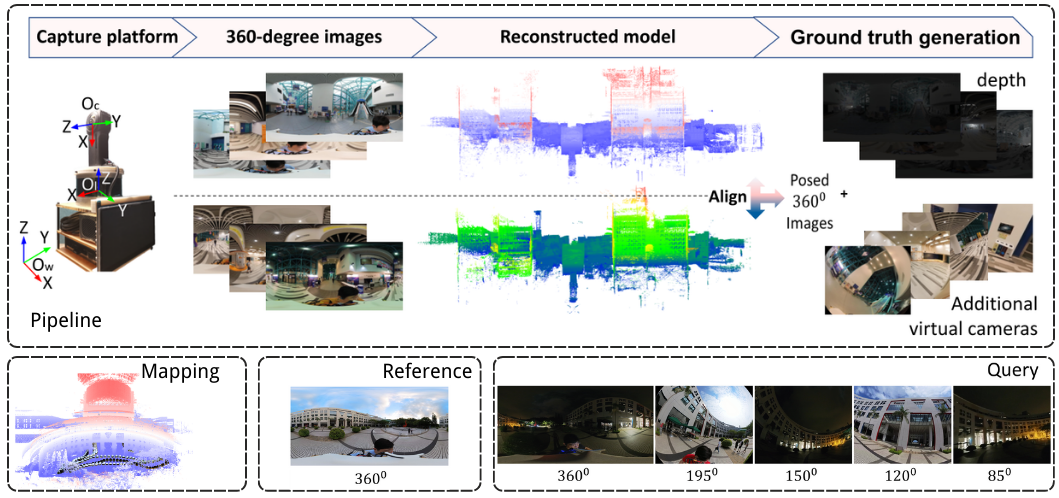}
    \vspace{-0.3cm}
    \caption{Overview of dataset collection and ground truth generation: 1) Use the platform to collect 360$^\circ$ images and frame-by-frame point clouds. Obtain real-time camera poses; 2) Apply optimization methodology to achieve data registration, resulting in a globally reconstructed point cloud model. Then, align the models in daytime and nighttime to get consistent poses; 3) Perform cropping to get \textit{\textit{virtual camera}} images and generate corresponding depth images. As a result, 360Loc takes advantage of 360$^\circ$ images for efficient mapping while providing query images in five different camera models in order to analyze the challenge of cross-domain visual localization.}
    \label{fig:pipeline}
\end{figure*}

Visual localization refers to predicting the 6DoF absolute pose (translation and rotation) of query images in a known scene. Accurate visual localization has wide applications in augmented reality (AR), navigation, and robotics. Over the last decade, many visual localization methods have been proposed, including feature matching-based approaches~\cite{dusmanu2019d2,sarlin2019coarse,taira2018inloc,sattler2016efficient, liu2017efficient}, scene coordinate regression~\cite{brachmann2021dsacstar,brachmann2017dsac,brachmann2018lessmore} and absolute pose regressors (APRs)~\cite{kendall2015posenet,kendall2017geometric,shavit2021learning}. Much of this progress has been driven by the availability of numerous datasets and benchmarks targeting different challenges, as shown in Table~\ref{tab:dsoverview}. However, existing methods and datasets focus on localization and mapping using pinhole images. Although the merits of 360$^\circ$ camera on visual perception have been recognized~\cite{huang2023360vot, PANDORA, osv2019}, the application of 360$^\circ$ cameras for visual localization is still under-explored. Recently, SensLoc~\cite{yan2023long} started to apply 360$^\circ$ cameras to facilitate data collection, but their pipeline cannot perform omnidirectional localization directly from the 360$^\circ$ images. 

This paper introduces 360Loc, a new challenging benchmark dataset to facilitate research on omnidirectional visual localization. The dataset contains 360$^\circ$ images captured in diverse campus-scale indoor and outdoor environments, featuring highly symmetrical and repetitive features, as well as interference of dynamic objects.
To capture this dataset,  we present a practical pipeline using a portable 360-camera-lidar platform to obtain reliable pose estimations of 360$^\circ$ cameras as ground truth.
Although 360$^\circ$  cameras present significant advantages for capturing reference data, real-life applications applying visual localization often rely on traditional cameras. Examples include robots equipped with fisheye cameras and phone-based AR applications using the embedded pinhole camera. 
This raises the problem of cross-device visual localization on image databases captured with 360$^\circ$ cameras. 
We thus supplement the reference database composed of 360$^\circ$ images with query frames including pinhole, fisheye and 360$^\circ$ cameras. 

We introduce the concept of \textit{virtual camera} to generate high-quality lower-FoV images with different camera parameters from 360$^\circ$ images.  
This enables a fair comparison of performance among queries from different devices in cross-device visual localization.
We adapt existing feature-matching-based methods and APRs to support 360$^\circ$ image queries and benchmark these methods for 360-based cross-device visual localization. 
Since different cameras present different imaging patterns, the cross-device domain gap is expected to lead to performance loss.
We extend the \textit{virtual camera} approach to data augmentation for end-to-end solutions such as image retrieval (IR) and APRs.

By conducting exhaustive evaluations, we demonstrate the advantages of 360$^\circ$ cameras in reducing ambiguity in visual localization on scenes featuring symmetric or repetitive features. We also show improvements against state-of-the-art (SOTA) baselines using the \textit{virtual camera} method for cross-device visual localization on images databases captured with 360$^\circ$ cameras.
These results provide novel insights on mapping using 360$^\circ$ images, enhancing the anti-ambiguity capability of query images, reducing domain gap cross-device in visual localization, and improving the generalization ability of APRs by applying \textit{virtual cameras}.

Our contribution can be summarized as follows:
\begin{itemize}
\item We propose a practical implementation of 360$^\circ$ mapping combining lidar data with 360$^\circ$ images for establishing the ground truth 6DoF poses.
    \item A virtual camera approach to generate high-quality lower-FoV images with different camera parameters from 360$^\circ$ views.
    \item A novel dataset for cross-device visual localization based on 360$^\circ$ reference images with pinhole, fisheye, and 360$^\circ$ query images.
    \item Demonstration of our approach's efficacy over state-of-the-art solutions for visual localization using 360$^\circ$ image databases, resulting in decreased localization ambiguity, reduced cross-device domain gap, and improved generalization ability of APRs.
\end{itemize}

\section{Related work}
\subsection{Visual Localization}
\textbf{Structure-based methods} predict camera poses by establishing 2D-3D correspondences  indirectly with local feature extractors and matchers~\cite{sarlin2019coarse,lowe2004distinctive,sarlin2020superglue,detone2018superpoint,sun2021loftr,tyszkiewicz2020disk} or directly with scene coordinate regression~\cite{brachmann2017dsac,brachmann2021dsacstar,brachmann2018lessmore}. HLoc~\cite{sarlin2019coarse,sarlin2020superglue} pipeline scales up to large scenes using image retrieval~\cite{arandjelovic2016netvlad, Berton_CVPR_2022_CosPlace, ge2020self, GARL17} as an intermediate step, which achieves SOTA accuracy on many benchmarks. This type of approach usually supports pinhole cameras with different intrinsic parameters. However, the performance of 360$^\circ$ and fisheye cameras has not been evaluated before due to the lack of support for 360$^\circ$ cameras in the Structure from Motion (SfM) tools like COLMAP~\cite{sattler2016efficient} and the lack of datasets for fisheye and 360$^\circ$ cameras.
~\cite{kim2021piccolo,kim2022cpo,kim2023calibrating} are point-cloud-based panorama localization methods for 360$^\circ$ queries but they do not consider cross-device visual localization.\\
\textbf{Absolute Pose Regressors} (APRs) are end-to-end learning-based methods that directly regress the absolute camera pose from input images without the knowledge of 3D models and establish 2D-3D correspondences. APRs~\cite{kendall2015posenet,kendall2017geometric,blanton2020extending,melekhov2017image,wu2017delving,shavit2021learning,chen2021direct,chen2022dfnet,brahmbhatt2018geometry,moreau2022coordinet} 
provide faster inference than structure-based methods at the cost of accuracy and robustness~\cite{sattler2019understanding}. Besides, APRs have generally only been tested on the ~\cite{bui2020eccv}, 7Scenes~\cite{shotton2013scene}, and Cambridge Landmarks~\cite{kendall2015posenet} datasets in previous studies. A notable characteristic of these datasets is that the training set and test set images were taken from the same camera. In this paper, we enhance cross-device pose regression for APRs by introducing virtual cameras as a data augmentation technique.

\begin{table*}
\centering
\setlength{\tabcolsep}{5pt} 
\resizebox{2.05\columnwidth}{!}{
\begin{tabular}{c|c|c|c|c|c}
\toprule 
Dataset & Scale and Environment & Challenges & Reference/Query type & Groundtruth Solution & Accuracy \\
\midrule
7Scenes~\cite{shotton2013scene} & Small Indoor & None & pinhole / pinhole & RGB-D & $\approx$ cm \\

RIO10~\cite{wald2020beyond} & Small Indoor & Changes & pinhole / pinhole &  VIO & $>d$ m \\

Baidu Mall~\cite{sun2017dataset} & Medium Indoor  & People, Ambiguous & pinhole / pinhole & lidar+Manual  & $\approx d$ m \\

Naver Labs~\cite{lee2021large} & Medium Indoor  & People, Changes & pinhole / pinhole & lidar+SfM & $\approx d$ m \\

InLoc~\cite{taira2018inloc} & Medium Indoor  & None & pinhole / pinhole & lidar+Manual &  $>d$ m \\

AmbiguousLoc~\cite{bui2020eccv} & Small Indoor  & Ambiguous & pinhole / pinhole & SLAM & $\approx$ cm \\

Achen~\cite{sattler2018benchmarking} & Large outdoor & People, Day-Night & pinhole / pinhole & SfM & $>d$ m \\

Cambridge~\cite{kendall2015posenet} & Medium outdoor & People, Weather & pinhole / pinhole & SfM & $>d$ m \\

San Francisco~\cite{chen2011city} & Large outdoor  & People, Construction & pinhole / pinhole & SfM+GPS & $\approx m$ \\

NCLT~\cite{carlevaris2016university} & Medium Outdoor + Indoor & Weather & pinhole / pinhole & GPS+SLAM+lidar & $\approx d$ m \\

ADVIO~\cite{cortes2018advio} & Medium Outdoor+Indoor  & People & pinhole / pinhole & VIO+Manual & $\approx m$ \\

ETH3D~\cite{schops2017multi} & Medium Outdoor + Indoor  & None & pinhole / pinhole & lidar+Manual & $\approx$ mm \\

 LaMAR~\cite{sarlin2022lamar} & Medium Outdoor+Indoor & \makecell{\footnotesize People, Weather, Day-Night, Construction, Changes, Ambiguous} & pinhole / pinhole & lidar+SfM+VIO & $\approx$ cm \\

SensLoc~\cite{yan2023long} & Large Outdoor & \makecell{\footnotesize People, Weather, Day-Night, Construction, Changes} & pinhole / pinhole & SL+VIO+RTK+Gravity &  $<$ dm \\


\midrule
\textbf{360Loc (ours)} & Medium Outdoor+Indoor & \makecell{\footnotesize People, Weather, Day-Night, Construction, Changes, \\ \textbf{Ambiguous}} & \makecell{\textbf{360 / (360 +} \\\textbf{pinhole + fisheye)}} & lidar+VIO & $\approx$ cm \\

\bottomrule
\end{tabular}
}
\vspace{-0.2cm}
\caption{Overview of popular visual localization datasets. No dataset, besides ours, consider 360$^\circ$ images as reference and query frames from pinhole, ultra-wide FoV fisheye, and 360$^\circ$ cameras.}
\label{tab:dsoverview}
\end{table*}

\subsection{Datasets}
The existing dataset has the following limitations. 
1). Most datasets~\cite{shotton2013scene,kendall2015posenet,wald2020beyond,taira2018inloc,bui2020eccv, carlevaris2016university} do not consider the need for cross-device localization, i.e., query images come from the same camera. Even though some datasets~\cite{sun2017dataset,cortes2018advio,sarlin2022lamar,sattler2018benchmarking,lee2021large,chen2011city, yan2023long, schops2017multi} take into account cross-device localization, these devices are only pinhole cameras with different camera intrinsic parameters and do not have particularly large domain-gaps. Compared to~\cite{liu2023low}, our pinhole and fisheye images are extracted from 360$^\circ$ images via virtual cameras, which makes less demands on the device and allows for a fair and more flexible comparison of the effects of different FoVs.
In this paper, our 360Loc datasets provide five kinds of queries from pinhole, fisheye and 360$^\circ$ cameras to promote the research of cross-device localization. 2). Now there is no 6DoF visual localization dataset and benchmark considering 360$^\circ$ reference images and 360$^\circ$ query images, even though \cite{armeni2017joint, kim2021piccolo,murrugarra2022pose} contain 360$^\circ$ images with 6DoF pose labels, they are not standard visual localization datasets with independent mapping/reference sequences and query sequences like datasets in Table~\ref{tab:dsoverview}. Other datasets~\cite{chen2011city,yan2023long} use 360$^\circ$ cameras for data collection, in the end they cropped 360$^\circ$ to perspective images and then tailor these images to the classical visual localization pipeline of pinhole cameras. 
The academic community is mainly driven by benchmarks where all training, reference, and query images are pinhole images
because they rely on SfM tools~\cite{sattler2016efficient} which does not support 360$^\circ$ cameras to obtain ground-truth (GT) and get sparse 3D point cloud models for recovering camera poses. However, we note that the 360$^\circ$ camera can cover the scene with greater efficiency than normal pinhole cameras with narrow Field-of-View (FoV), which makes 360$^\circ$ images particularly suitable as reference images. 3) Although the current dataset has explored the challenges of visual localization from various aspects such as weather variations, day-night transitions, scene changes, and moving individuals and objects~\cite{wald2020beyond,kendall2015posenet,yan2023long,sarlin2022lamar,sattler2018benchmarking,lee2021large}, there is still insufficient research specifically targeting highly ambiguous environments which contain symmetries, repetitive structures and insufficient textures. Only two indoor datasets~\cite{bui2020eccv, sun2017dataset} and LaMAR~\cite{sarlin2022lamar} consider challenges in ambiguous environments. In this paper, we studied 4 ambiguous scenes from both indoor and outdoor environments with a scale much larger than dataset~\cite{bui2020eccv} (See Figure~\ref{fig:overview}). We conduct exhaustive assessments of image retrieval, local matching localization, and absolute pose regression to show that queries from the $360^\circ$ camera are harder to obtain plausible solutions than other queries from cameras with narrower FoV.

\section{The 360Loc Dataset}
\begin{figure}
    \centering
    \subfloat[Atrium]{\includegraphics[width=0.40\linewidth]{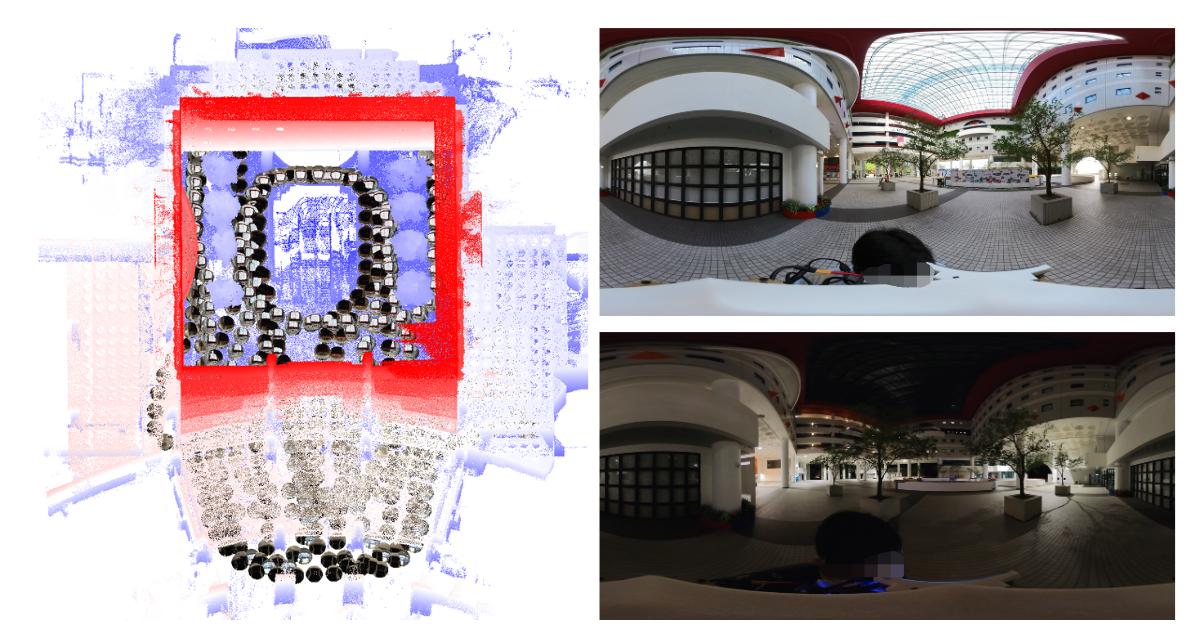}}
    \subfloat[Concourse]{\includegraphics[width=0.59\linewidth]{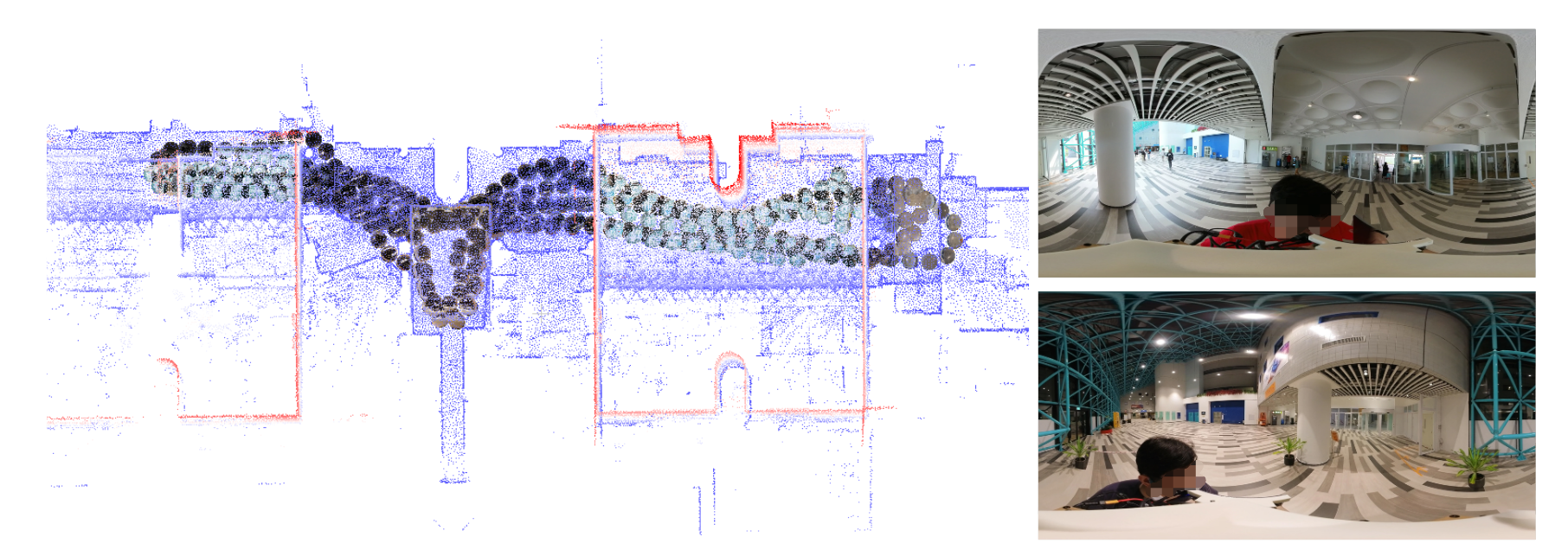}}\\
    \subfloat[Piatrium]
    {\includegraphics[width=0.40\linewidth]{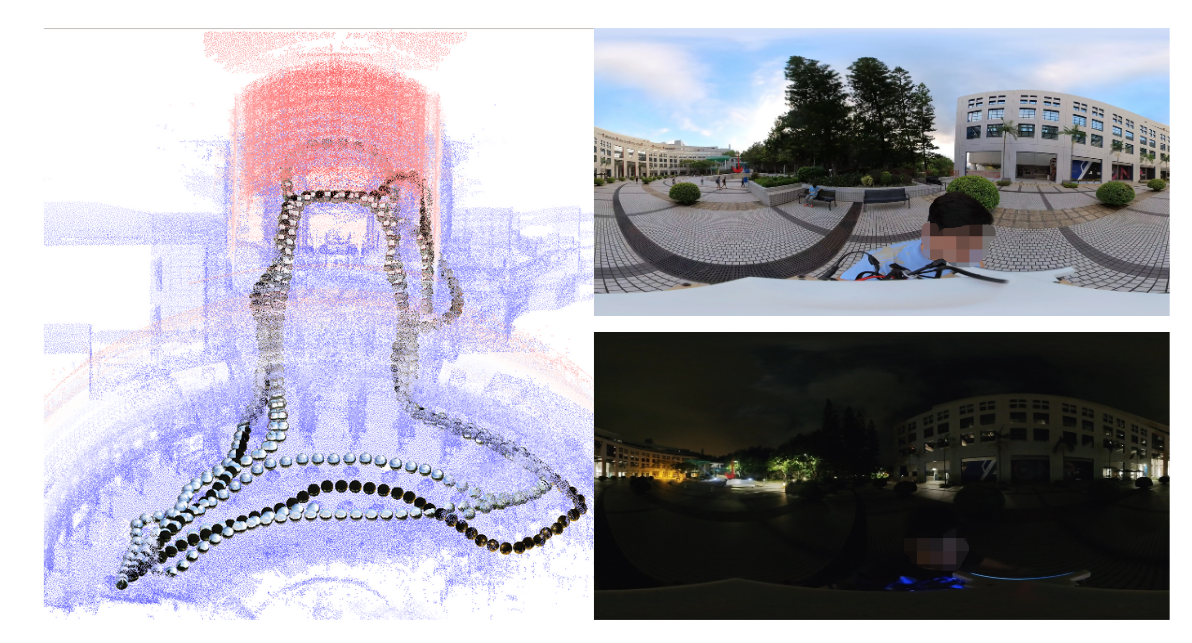}}
    \subfloat[Hall]{\includegraphics[width=0.59\linewidth]{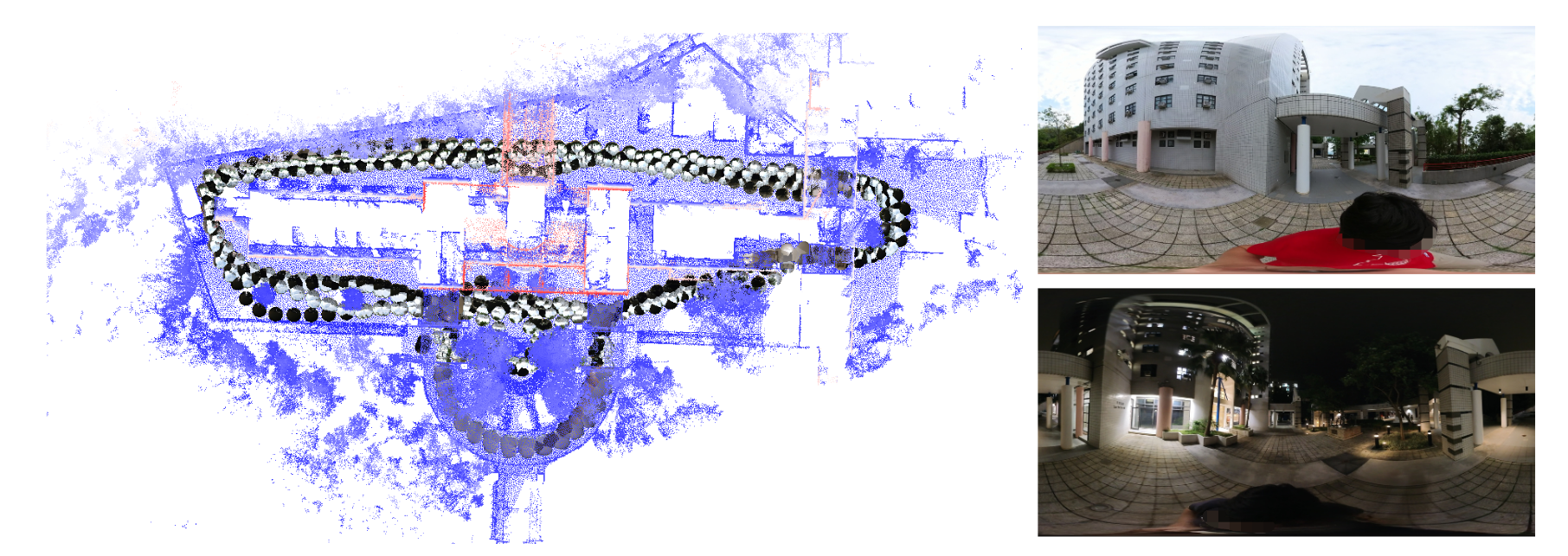}}
    \caption{The four scenes in 360Loc, all four scenes contain symmetrical, repetitive structures and moving objects. The camera trajectories are visualized as spheres. }
    \label{fig:overview}
\end{figure}

The 360Loc dataset contains 4 locations from a local university. Figure~\ref{fig:overview} displays the reference point cloud and example frames from each scene. Atrium is inside a building with a surrounding structure that exhibits a high degree of symmetry and repetition, making it a highly ambiguous environment. Concourse is a large indoor scene with many moving people, which can be used for evaluating the robustness of any localization algorithm in scenes with many moving objects. Piatrium is a scene containing both indoor Atrium and outdoor environments, covering an outdoor piazza with coffee shops, bookstores, and souvenir shops. Hall is a modern building of a student dormitory.

\begin{table}
\centering
\setlength{\tabcolsep}{1pt} 
{\footnotesize 
\begin{tabular}{c|c|c|c|c}
\toprule
Symbol & Name & Field of View & Resolution & Type \\
\midrule 
$c_0$ & 360 & 360$^\circ$ & 6144$\times$3072 & reference/query \\
$c_1$ & fisheye1 & 120$^\circ$ & 1280$\times$1024 & query \\
$c_2$ & fisheye2 & 150$^\circ$ & 1280$\times$1024 & query \\
$c_3$& fisheye3 & 195$^\circ$ & 1280$\times$1024 & query \\
$c_4$ & pinhole & 85$^\circ$ & 1920$\times$1200 & query \\
\bottomrule
\end{tabular}
}
\vspace{-0.2cm}
\caption{The representation and parameters of 5 cameras.}
\label{tab:caminfo}
\end{table}

\subsection{360 Mapping Platform}
We utilized the handheld multimodal data acquisition platform depicted in Figure~\ref{fig:pipeline} for data collection. This platform incorporates a 360$^\circ$ camera, a Velodyne VLP-16 multi-line lidar, an NUC mini-computer, and a display screen. 
Figure~\ref{fig:pipeline} also illustrates the relative relationship among the 360$^\circ$ camera coordinate system $\mathbf{O_c\text{-}XYZ}$, the lidar coordinate system $\mathbf{O_l\text{-}XYZ}$ as well as the world coordinate $\mathbf{O_w\text{-}XYZ}$.
The portable 360 camera equipped on this device can capture high-resolution omnidirectional images with a resolution of 6144 x 3072 (2:1 aspect ratio). It also features a built-in six-axis gyroscope that provides stabilization support, making it suitable for handheld mobile data capture. 
The Velodyne VLP-16 multi-line lidar has a FoV of 360$^\circ$$\times$30$^\circ$, angular resolution of 0.2$^\circ$$\times$2.0$^\circ$, and rotation rate of 10Hz, offering a comprehensive 360$^\circ$ environmental view.
Regarding the calibration of the extrinsic poses between the lidar and the 360$^\circ$ camera, we employed a calibration toolbox~\cite{koide2023general} that applies to both lidar and camera projection models. This toolbox utilizes the SuperGlue~\cite{sarlin2020superglue} image matching pipeline to establish 2D-3D correspondences between the lidar and camera image. 
We perform pseudo-registration by synchronizing the two data modalities, images, and point clouds. Eventually, we use graph-based SLAM techniques for continuous pose estimations.
In the four scenes, a total of 18 independent sequences of 360$^\circ$ images were captured (12 daytime, and 6 nighttime), resulting in a total number of 9334 images. For each scene, we selected a specific sequence captured during the daytime as the reference images, while the remaining images were defined as query images of the 360$^\circ$ image type. We provide more details and show why 360$^\circ$ mapping is superior to pinhole SfM in ambiguous scenes with repetitive and symmetric structures in the supplementary material.

\begin{table}[t]
\centering
\setlength{\tabcolsep}{1pt} 
\resizebox{\columnwidth}{!}{
\begin{tabular}{l|c|ccccc|c} 
\toprule
& \multicolumn{1}{c}{\# Frames}& \multicolumn{5}{|c|}{\# Frames Query (day / night)}& Spatial \\
Scene & Reference 360  & 360  & Pinhole  & Fisheye1  & Fisheye2  & Fisheye3  & Extent $(\mathrm{m})$\\
\midrule 
Concourse & 491 & 593/514 & 1186/1028  &  1186/1028  & 1186/1028  & 1186/1028   & $93\times15  $ \\
Hall & 540 & 1123/1061 &  2246/2122 &  2246/2122 &2246/2122  & 2246/2122  & $ 105\times 52  $ \\
Atrium & 581 & 875/1219 & 1750/2438  & 1750/2438  & 1750/2438 & 1750/2438  & $65\times 36  $ \\
Piatrium & 632 & 1008/697 & 2016/1394  & 2016/1394  & 2016/1394 & 2016/1394  & $98\times70  $ \\
\bottomrule
\end{tabular}%
}
\vspace{-0.2cm}
\caption{360Loc dataset description.}
\label{tab:dsdesc}
\end{table}

\begin{figure}
    \centering
    \includegraphics[width=0.4\textwidth]{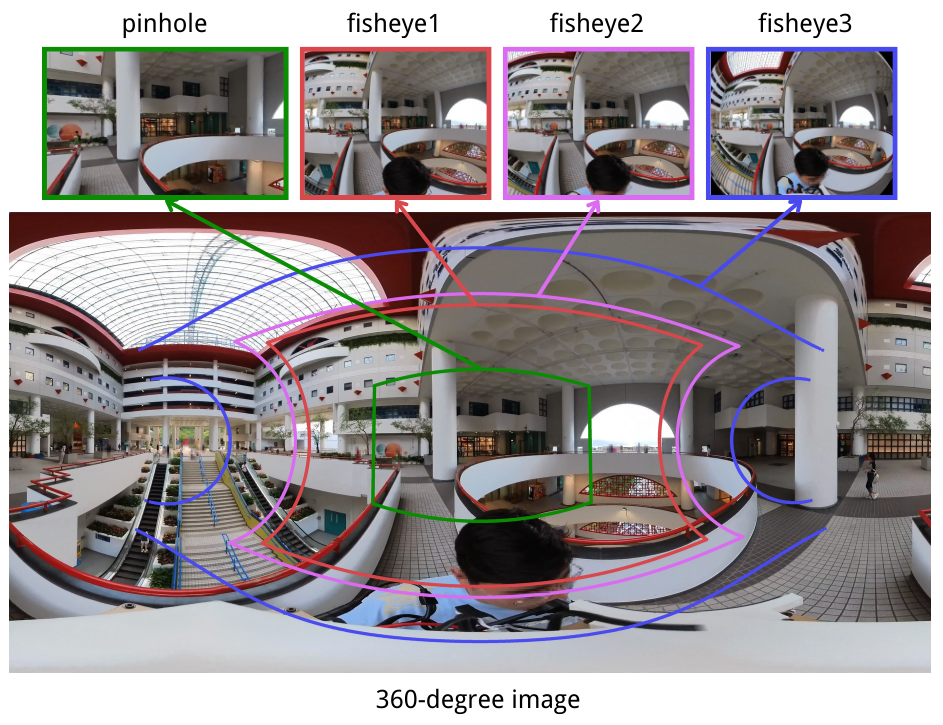}
    \vspace{-0.3cm}
    \caption{Illustration of obtaining virtual camera images through random poses and image cropping.}
    \label{fig:virtualcam}
\end{figure}

\subsubsection{Cross-device Queries}
\label{subsec:cross-device}
To enable a rigorous comparison of the difference in the performance of different FoV queries for visual localization tasks, we created four virtual cameras with diverse FoV from 360$^\circ$ cameras, which are shown in Figure~\ref{fig:overview}. 
Given a 360$^\circ$ image $\mathcal{I}_{c_0}$, the corresponding virtual camera with preconfigured intrinsic parameters is extracted by 
\begin{equation}
\mathcal{I}_{c_n} = \Psi_{c_n}(\mathcal{I}_{c_0}) = \pi_{c_n}^{-1} \left( \pi_{c_0}\left( \bm{R} \mathcal{I}_{c_0}\right)  \right),
\label{eq:getvircam}
\end{equation}
where $\pi_{c_n}$ denote the projection function of virtual camera and $\pi_{c_0}$ is the projection function of 360$^\circ$ camera. $\bm{R}\in SO\left(3\right)$ is a random relative rotation matrix to increase the diversity of views representing the scenes. 
Moreover, the inversed operation $\Psi_{c_n}^{-1}$ can convert the ${c_n}$ image back to a 360$^\circ$ image.
As reported in Table~\ref{tab:caminfo}, the virtual cameras include an undistorted pinhole model with 85$^\circ$ FoV and three fisheye cameras in Dual Sphere mode ~\cite{usenko2018double} with 120$^\circ$, 150$^\circ$, and 195$^\circ$ FoV respectively. 
Table~\ref{tab:dsdesc} presents the number of image frames in the 360Loc dataset.



\subsection{Ground Truth Generation}
Besides the graph-based optimization in SLAM, we designed a set of offline optimization strategies to further improve the accuracy of camera pose estimation. 
After the acquisition of precise dense point cloud reconstructions and poses of 360$^\circ$ cameras, an Iterative Closest Point (ICP) algorithm is applied to align models between reference and the query sequences in the same scene. Moreover, we reconstructed the mesh model of the scenes and generated corresponding depth maps of 360$^\circ$ cameras.

\textbf{Bundle Adjustment (BA) of lidar mapping.} Incremental map construction can suffer from accumulating errors due to environmental degradation. We utilized a BA framework based on feature points extracted from lidar to refine the map and the poses. The optimization process involved minimizing the covariance matrix to constrain the distances between feature points and edge lines or plane features that are mutually matched. 

First, we utilize an octree data structure to perform adaptive voxelization-based feature extraction. In this method, the point cloud map is segmented into voxels of predetermined size. Each voxel is checked to determine if its points ${\bm{P}}_{u}^f$ lie on a plane or a line, where $u\in \{1,2,\dots,U \}$, obtained from the $u$-th frame of lidar scans. If not, the voxel is recursively subdivided using an octree structure until each voxel contains points ${\bm{P}}_{u}^f$ belonging to the same feature. Let's assume that the pose of the lidar in each frame is $\bm{\eta} = \{\bm{\eta}_1,\bm{\eta}_2,\dots,\bm{\eta}_M\}$, where $\bm{\eta}_u = (\bm{R}_u,\bm{t}_u|\bm{R}_u\in SO\left(3\right), \bm{t}_u \in \mathbb{R}^3)$. In that case, 
the feature points in the global map can be represented as follows:
\begin{equation}
\bm{P}_u=\bm{R}_{u}\times{\bm{P}}_{u}^f+\bm{t}_{u}.
\label{eq:globpts}
\end{equation}
After simplifying the lidar map to edge or plane features, the process of BA becomes focused on determining the pose $\bm{\eta}$ and the location of the single feature, which can be represented as $\left(\bm{n}_f,\bm{q}\right)$, where $\bm{q}$ represents the location of a specific feature, $\bm{n}_f$ is the direction vector of an edge line or the normal vector of a plane. 
To minimize the distance between each feature point and the corresponding feature, we can utilize the BA:
\begin{equation}
\left(\bm{\eta}^*,\bm{n}_f^*,\bm{q}^*\right)=\mathop{\arg\min}\limits_{\bm{\eta},\bm{n}_f,\bm{q}}\frac{1}{U}\sum\limits_{u=1}^{U}\left(\bm{n}_f^T\left(\bm{P_u}-\bm{q}\right)\right)^2.
\label{eq:BA}
\end{equation}
It has been proved that when the plane's normal vector is set to the minimum eigenvector, and $\bm{q}$ is set to the centroid of the feature, i.e. $\bm{q}=\hat{\bm{P}}=\frac{1}{U}\sum\limits_{u=1}^{U}\bm{P}_u$, Eq.~\ref{eq:BA} reaches its minimum value. Additionally, the BA problem in lidar mapping has a closed-form solution that is independent of the features $\left(\bm{n}_f, \bm{q}\right)$~\cite{liu2021balm}. It can be simplified to the following problem:
\begin{equation}
\bm{\eta}^* = \mathop{\arg\min}\limits_{\bm{\eta}}\lambda_{\min}\left(\bm{A}\right),
\label{eq:BAsol}
\end{equation}
where, $\lambda$ represents the eigenvalue of $A$, and
\begin{equation}
\bm{A} = \frac{1}{U}\sum\limits_{u=1}^{U}\left(\bm{P}_u-\hat{\bm{P}}\right)\left(\bm{P}_u-\hat{\bm{P}}\right)^T.
\label{eq:BAsimp}
\end{equation}
Now, the BA problem is simplified by adjusting the lidar pose $\bm{\eta}$ to minimize the smallest eigenvalue $\lambda_3$ of the point covariance matrix $\mathbf{A}$ defined in Eq.~\ref{eq:BAsimp}. By employing this strategy, we refined the pose $\bm{\eta}$ of each frame and the edge or plane features in the lidar map.

\textbf{Refined cameras poses.} 
The poses of 360$^\circ$ camera obtained from online SLAM are further optimized by the registration with respect to the dense refined point cloud model.
Taking the pre-calibrated extrinsic parameters as the initial guess, we used the RANSAC to refine the lidar-camera transformation~\cite{koide2023general}. This registration process is based on the normalized information distance (NID)~\cite{stewart2014localisation}, which serves as a mutual information-based cross-modal distance metric.  
Finally, we align the reference models and query models into the same coordinate system to generate the ground truth for the query sequences.
Specifically, we utilize the CloudCompare toolbox~\cite{girardeau2016cloudcompare} to manually select feature points across multiple point cloud models as initial values. Then, we employ the ICP algorithm to register the point cloud models together. Afterwards, we employed a practical approach to volumetric surface reconstruction called Truncated Signed Distance Functions (TSDFs)~\cite{vizzo2022vdbfusion} to achieve the reconstruction from point clouds to meshes with an efficient and sparse data structure called Voxel Data Base (VDB)~\cite{museth2013vdb}. At this stage, we can utilize the ray-mesh intersection method~\cite{trimesh} to cast rays from cameras onto the mesh model. By intersecting the rays with the mesh, we can determine the depths of the corresponding points on the mesh surface.
\begin{figure}
    \centering
    \includegraphics[width=0.47\textwidth]{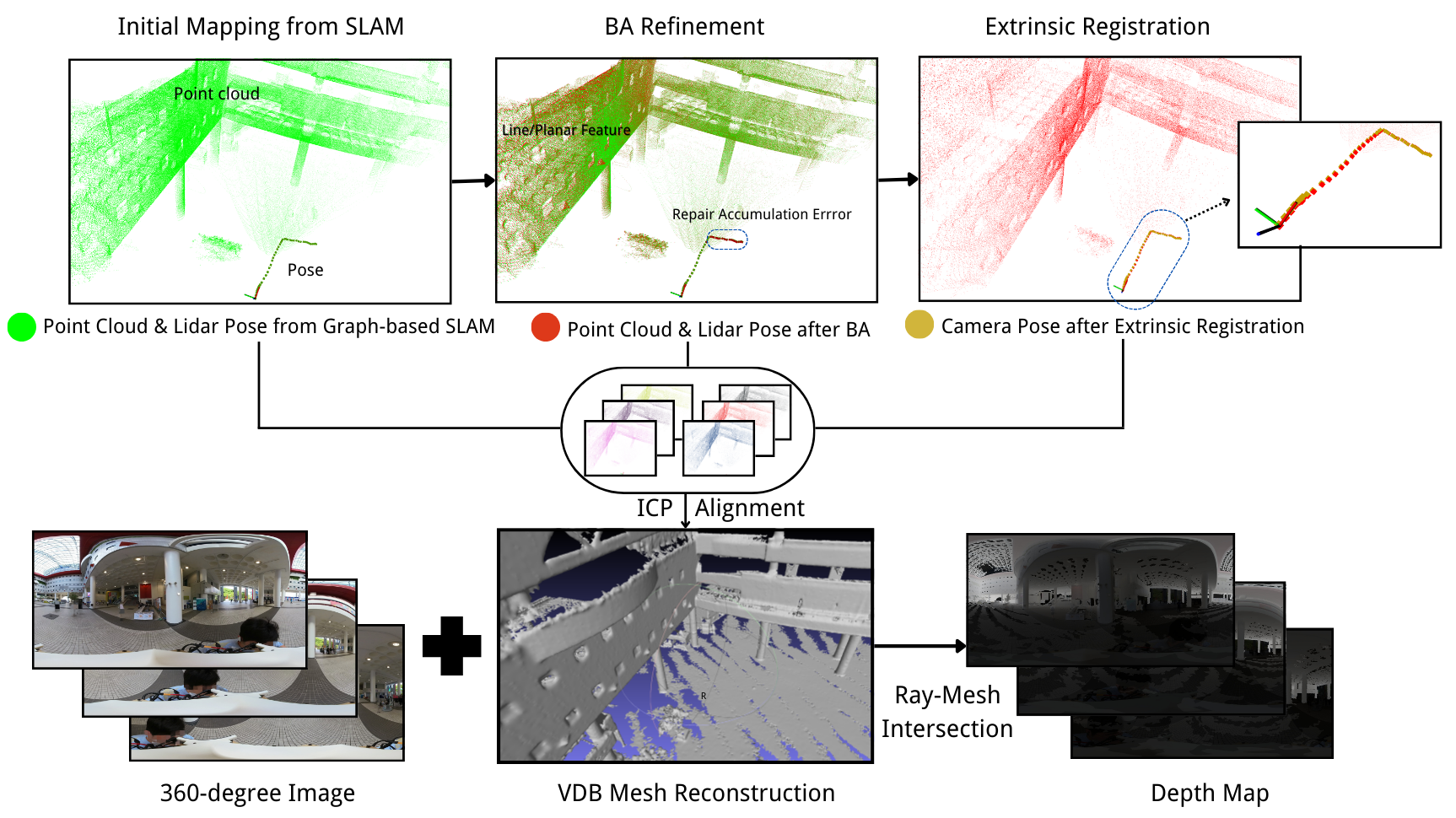}
    \caption{Overview of GT generation.}
    \label{fig:GT}
\end{figure}
After a series of joint optimizations between multiple modalities, we have generated a set of GT data. Figure~\ref{fig:overview} shows some instances. This GT data includes reference images $\mathcal{I}^r_{c_0}$, the depth maps $D^r_{c_0}$ of the reference images, and the reference maps containing the point cloud models $\mathcal{P}$, mesh models $M$, as well as camera pose odometry $\{\bm{\xi}\}$. Figure~\ref{fig:GT} summarizes the GT generation.


\section{Omnidirectional Visual Localization}
We extend the current feature-matching-based and absolute pose regression pipelines for omnidirectional visual localization.
Given a query image $\mathcal{I}^q$ in any camera model, we seek to estimate its poses within the environment modeled by 360$^\circ$ images $\textbf{I}^r$. To minimize the domain gap between the query image from $c_1$, $c_2$, $c_3$,$c_4$ and reference images, we explore virtual cameras (VC) in two ways: VC1, remapping query images to 360 domain using $\Psi_{c_n}^{-1}$; VC2, rectifying 360$^\circ$ images into queries' domains using $\Psi_{c_n}$. 


\subsection{Feature-matching-based Localization}

Most feature-matching-based techniques first perform IR to reduce the search space before estimating the pose.

\subsubsection{Image Retrieval}
\label{subsec:ir}

For method VC1, if query $\mathcal{I}^q$ captured from $c_0$, we retrieve the
$k$ most similar images from $\textbf{I}^r$ by calculating and sorting $simi_{\cos}(\mathcal{F}(\mathcal{I}^q),\mathcal{F}(\mathcal{I}^r))$, $\mathcal{I}^r \in \textbf{I}^r $ and $\mathcal{F}(\cdot)$ denotes the function to map each image to the global feature domain. $simi_{\cos}(\cdot)$ is cosine similarity for two feature embeddings. If query $\mathcal{I}^q$ captured from $c_1,c_2,c_3,c_4$, we then retrieve top-$k$ reference images based on $simi_{\cos}(\mathcal{F}(\Psi_{c_n}^{-1}(\mathcal{I}^q)),\mathcal{F}(\mathcal{I}^r))$, $\mathcal{I}^r \in \textbf{I}^r$.



In method VC2, we expand the global features for each 360$^\circ$ reference image by cameras $c$ including virtual pinhole cameras forming a cube map and virtual fisheye cameras.
We define the similarity score between $\mathcal{I}^q$ and $\mathcal{I}^r$ as:
\begin{equation}
    \max( simi_{\cos} (\mathcal{F}(\mathcal{I}^q), \mathcal{G}_{\mathcal{F}}(\mathcal{I}^r)) ,
\label{eq:ir}
\end{equation}
where global feature group of reference is $\mathcal{G}_{\mathcal{F}}(\mathcal{I}^r) = \{\mathcal{F}(\Psi_c(\mathcal{I}^r)), \dots\}$.
 We use the highest similarity value calculated from $\mathcal{F}(\mathcal{I}^q)$ and $\mathcal{G}_{\mathcal{F}}(\mathcal{I}^r)$ as the similarity score for each $\mathcal{I}^r$ to ensure retrieve
$k$ most similar 360$^\circ$ reference images
because some rectified images are from the same $\mathcal{I}^r$.  Note that we can eliminate the domain gap during the image retrieval step in this way. 



\subsubsection{Local Feature Matching and Pose Estimation}
For each pinhole query frame, we retrieve relevant reference images, match their local features, leverage the depth maps $D_{c_0}$ to establish the 2D-3D correspondences, and finally estimate a pose with PnP+RANSAC. Unlike ~\cite{yan2023long,chen2011city}, we directly match query image with retrieved 360$^\circ$ reference images described in Section~\ref{subsec:ir}. For query images from $c_0,c_1,c_2,c_3$, i.e., fisheye and 360$^\circ$ query frames, we utilize the function that calculates pose error in sphere camera model in OpenGV~\cite{kneip2014opengv} library for PnP+RANSAC.


\subsection{Absolute Pose Regression}
APRs train deep neural networks to regress the 6DoF camera pose of a query image. 

\noindent\textbf{PN}. PoseNet (PN) is the first APR model. Since there is no open source code~\cite{kendall2015posenet,kendall2017geometric}, we follow the modification in~\cite{brahmbhatt2018geometry,melekhov2017image} and use ResNet34~\cite{he2016deep} as the backbone network. 


\noindent\textbf{MS-T}. MS-Transformer~\cite{shavit2021learning} is an APR model incorporating attention and implementing transformers as backbone. 
\textbf{We note APR methods using our virtual camera method, VC2, as APR$^{vc2}$}. The difference between APR and APR$^{vc2}$ is the training stage. For APR baselines, the training set is $\textbf{I}^r$. For APR$^{vc2}$, they are trained with 360$^\circ$ images, cropped pinhole images, and cropped fisheye images, i.e., $\textbf{I}^r\cup \Psi_c(\textbf{I}^r)$ introduced in Section~\ref{subsec:ir} and Eq.~\ref{eq:getvircam}. 

All APR models are implemented in Python using PyTorch~\cite{paszke2019pytorch}. During training, all input images are resized to $256\times 256$ and then randomly cropped to $224\times 224$. For both PN and MS-T, we set an initial learning rate of $\lambda= 10^{-4}$ and a batch size of 32 for 300 epochs of each scene. Training and evaluation in Section~\ref{sec:exp} are performed on an NVIDIA GeForce GTX 3090 GPU.

\begin{table*}[]
    \centering
    \footnotesize
    \setlength{\tabcolsep}{2pt} 
    \resizebox{2\columnwidth}{!}{
        \begin{tabular}{c|ccccc|ccccc|ccccc|ccccc}
            \toprule
            \multicolumn{1}{c}{} &\multicolumn{5}{c}{{NetVLAD~\cite{arandjelovic2016netvlad}}}&\multicolumn{5}{c}{{Cosplace~\cite{Berton_CVPR_2022_CosPlace}}}&\multicolumn{5}{c}{{OpenIBL~\cite{ge2020self}}}&\multicolumn{5}{c}{{AP-GeM~\cite{GARL17}}}
            \\ \cmidrule(lr){2-6} \cmidrule(lr){7-11} \cmidrule(lr){12-16} \cmidrule(lr){17-21}
            Query & R@1 & R@5 &P@5& R@10& P@10& R@1 & R@5 &P@5& R@10& P@10& R@1 & R@5 &P@5& R@10& P@10& R@1 & R@5 &P@5& R@10& P@10\\
            \midrule 

            pinhole&0.23&0.45&0.22&0.58&0.22 &0.15&0.26&0.15&0.33&0.15  &0.18&0.36&0.18&0.48&0.18 &0.2&0.37&0.2&0.47&0.2\\
            +VC1&0.24&0.45&0.24&0.57&0.23    &0.21&0.33&0.21&0.41&0.21  &0.21&0.39&0.21&0.5&0.2  &0.25&0.42&0.25&0.53&0.24\\
            +VC2&\marksecond{0.5}&\marksecond{0.67}&\marksecond{0.48}&\marksecond{0.75}&\marksecond{0.47}     &\marksecond{0.32}&\marksecond{0.41}&\marksecond{0.32}&\marksecond{0.48}&\marksecond{0.31}  &\marksecond{0.51}&\marksecond{0.67}&\marksecond{0.49}&\marksecond{0.75}&\marksecond{0.47} &\marksecond{0.5}&\marksecond{0.68}&\marksecond{0.49}&\marksecond{0.77}&\marksecond{0.47}\\
            \midrule
            fisheye1&0.42&0.67&0.41&0.77&0.39  &0.28&0.43&0.28&0.52&0.28 &0.37&0.58&0.36&0.69&0.34 &0.35&0.55&0.34&0.66&0.33\\
            +VC1&0.51&0.72&0.49&0.8&0.47        &0.36&0.48&0.35&0.56&0.34  &0.52&0.7&0.5&0.79&0.48 &0.43&0.62&0.42&0.72&0.4\\
            +VC2&\marksecond{0.73}&\marksecond{0.91}&\marksecond{0.63}&\marksecond{0.95}&\marksecond{0.57}     &\marksecond{0.63}&\marksecond{0.85}&\marksecond{0.51}&\marksecond{0.92}&\marksecond{0.43}  &\marksecond{0.74}&\marksecond{0.91}&\marksecond{0.62}&\marksecond{0.95}&\marksecond{0.54}  &\marksecond{0.65}&\marksecond{0.88}&\marksecond{0.57}&\markfirst{0.94}&\marksecond{0.51}\\
            \midrule
            fisheye2&0.45&0.7&0.44&0.8&0.42     &0.3&0.46&0.31&0.55&0.31   &0.41&0.62&0.4&0.73&0.38   &0.38&0.59&0.36&0.68&0.35\\
            +VC1&0.54&0.74&0.52&0.83&0.49       &0.37&0.49&0.36&0.57&0.35  &0.56&0.73&0.54&0.81&0.51  &0.46&0.65&0.45&0.74&0.43\\
            +VC2&\marksecond{0.74}&\marksecond{0.92}&\marksecond{0.65}&\marksecond{0.95}&\marksecond{0.58}       &\marksecond{0.64}&\marksecond{0.87}&\marksecond{0.53}&\marksecond{0.93}&\marksecond{0.45}  &\marksecond{0.76}&\marksecond{0.92}&\marksecond{0.65}&\markfirst{0.96}&\marksecond{0.56} &\marksecond{0.67}&\marksecond{0.89}&\marksecond{0.58}&\markfirst{0.94}&\marksecond{0.52}\\
            \midrule
            fisheye3&0.57&0.79&0.55&0.86&0.52   &0.4&0.56&0.4&0.65&0.4   &0.53&0.74&0.51&0.83&0.49 &0.45&0.66&0.43&0.75&0.41\\
            +VC1&0.63&0.81&0.61&0.88&0.58       &0.48&0.61&0.48&0.68&0.47  &0.67&0.82&0.65&0.88&\marksecond{0.61}  &0.55&0.73&0.53&0.81&0.51 \\
            +VC2&\marksecond{0.77}&\markfirst{0.93}&\marksecond{0.68}&\markfirst{0.96}&\marksecond{0.61} &\marksecond{0.69}&\marksecond{0.89}&\marksecond{0.58}&\marksecond{0.94}&\marksecond{0.5} &\marksecond{0.79}&\marksecond{0.93}&\marksecond{0.68}&\markfirst{0.96}&0.6 &\marksecond{0.67}&\markfirst{0.9}&\marksecond{0.59}&\markfirst{0.94}&\marksecond{0.54}\\
            \midrule
            360&\markfirst{0.79}&0.86&\markfirst{0.77}&0.88&\markfirst{0.73} &\markfirst{0.92}&\markfirst{0.95}&\markfirst{0.91}&\markfirst{0.96}&\markfirst{0.89}   &\markfirst{0.89}&\markfirst{0.94} &\markfirst{0.88}&0.95&\markfirst{0.83} &\markfirst{0.79}&\markfirst{0.9}&\markfirst{0.77}&\markfirst{0.94}&\markfirst{0.72}\\
            \bottomrule
        \end{tabular}
    }
    \caption{Image retrieval results based on 360$^\circ$ reference database average over four scenes, the recall, and precision for the top $k$ retrieved images, $k = 1,5,10$. \marksecond{\footnotesize $\#$} indicates the highest value  of R@$k$ and P@$k$ for each device w and w/o virtual cameras (VC1, VC2). Best results for all devices of R@$k$ and P@$k$ are in bold with \markfirst{\footnotesize $\#$}. }
    \label{tab:bs_ir}
\end{table*}

\begin{table*}[]
	\centering
	\scriptsize
	\tabcolsep=0.02cm 
	\resizebox{2.05\columnwidth}{!}{
		\begin{tabular}{c|cc |cc |cc |cc |cc |cc}
			\toprule
			\multicolumn{1}{c}{}&\multicolumn{6}{c}{NetVLAD~\cite{arandjelovic2016netvlad}}& \multicolumn{6}{c}{CosPlace~\cite{Berton_CVPR_2022_CosPlace}}\\\cmidrule(lr){2-7}\cmidrule(lr){8-13}
			\multicolumn{1}{c}{}& \multicolumn{2}{c}{\scriptsize DISK + LG} & \multicolumn{2}{c}{\scriptsize SP + LG} & \multicolumn{2}{c}{\scriptsize SP + SG } & \multicolumn{2}{c}{\scriptsize DISK + LG} & \multicolumn{2}{c}{\scriptsize SP + LG} & \multicolumn{2}{c}{\scriptsize SP + SG } \\\cmidrule(lr){2-3} \cmidrule(lr){4-5} \cmidrule(lr){6-7}  \cmidrule(lr){8-9} \cmidrule(lr){10-11} \cmidrule(lr){12-13} 
			\multicolumn{1}{c}{}& Day & Night  & Day & Night  & Day & Night  & Day & Night  & Day & Night & Day & Night \\
			\midrule
			pinhole     &6.0/11.3/24.6  & 1.7/4.4/10.3 & 8.0/14.9/30.9 & 2.2/5.5/13.5&8.4/15.2/30.7 &2.3/5.6/12.3& 4.2/7.8/18.0 &1.6/3.5/8.6 & 4.8/10.2/22.1 &1.9/4.7/11.1 & 5.4/10.4/21.1 &2.1/4.7/10.4 \\
			+VC1     &  8.5/14.0/23.5&2.2/4.1/7.9  & 10.4/17.0/27.5&2.9/5.3/10.1 &10.9/17.8/28.5&2.8/5.6/9.9 &  6.1/10.8/21.1&1.7/3.6/8.2 & 7.5/13.2/22.5&2.0/4.5/9.6 &7.6/13.5/22.8&2.1/4.7/9.6 \\
			+VC2     & \marksecond{14.2/22.2/35.5}& \marksecond{4.1/7.8/13.6} &  \markfirst{19.8}/\marksecond{29.7/42.9}&\marksecond{6.1/10.4/16.9} & \markfirst{21.6/33.2}/\marksecond{49.7}&\marksecond{5.9 / 11.0 / 18.4}& \marksecond{8.0/13.1/23.5}&\marksecond{2.5/4.6/9.1} &  \marksecond{10.7/16.4/26.6}&\marksecond{3.0/5.7/11.4} & \marksecond{11.6/18.5/30.5}&\marksecond{3.5/6.8/12.8}\\
			\midrule
			fisheye1     & 1.6/4.4/17.7&0.5/1.8/7.4  &  1.9/5.4/20.1&0.7/2.3/10.5 &  1.6/4.7/18.4&0.5/1.9/8.2& 0.8/2.5/11.8&0.4/1.4/5.8 &  1.0/3.5/13.0&0.5/1.4/8.2 & 0.9/3.4/12.1&0.3/1.4/7.0 \\
			+VC1    &  3.3/9.2/27.6&0.8/2.7/9.6  &  4.1/10.6/32.2&1.4/4.4/14.9 &  3.0/9.5/29.6&0.9/3.1/11.7& 2.3/5.5/19.4&0.5/1.6/7.3&  2.1/6.1/19.9&0.7/2.2/9.0 &  1.9/5.5/19.1&0.5/1.9/7.3 \\
			+VC2     &  \marksecond{3.9/10.5/33.0}& \marksecond{1.0/4.0/14.6} & \marksecond{4.3/12.4/38.2}&\marksecond{1.9/6.4/21.8} &  \marksecond{3.6/11.0/34.5}&\marksecond{1.1/5.3/19.4}&  \marksecond{2.5/6.9/25.3}&\marksecond{0.8/2.8/12.2} & \marksecond{2.8/8.2/29.0}&\marksecond{1.3/4.6/18.0} &  \marksecond{2.1/7.1/26.7}&\marksecond{1.0/4.0/16.2} \\
			\midrule
			fisheye2     &  1.6/4.9/20.9&0.5/2.0/8.7  &  1.9/6.7/23.2&0.8/3.0/11.8 & 1.7/5.2/19.5&0.7/2.5/9.9&  1.3/3.5/14.2&0.4/1.6/6.9 & 1.2/3.8/15.2&0.5/1.5/9.1 &  1.2/3.9/12.9&0.6/1.6/7.2 \\
			+VC1     &  \marksecond{4.3}/10.8/30.9&0.8/3.0/11.2 & 4.7/12.4/34.1&1.8/5.4/15.8 &  \marksecond{4.1}/10.6/31.5&1.1/3.6/13.7& 2.5/6.5/20.6&0.5/1.7/7.4 & 2.5/7.0/22.1&0.8/2.4/9.4 &2.2/6.8/20.2&0.5/2.1/8.0 \\
			+VC2    & \marksecond{4.3/11.0/34.4}&\marksecond{1.1/4.7/17.3}  &  \marksecond{5.1/14.0/41.1}&\marksecond{2.0/7.2/24.8} &  3.7/\marksecond{11.5/36.8}&\marksecond{1.5/5.9/21.2}& \marksecond{2.8/7.3/27.1}&\marksecond{0.8/2.9/13.4}&  \marksecond{2.9/8.9/32.0}&\marksecond{1.6/5.3/20.1 }&\marksecond{2.5/8.0/27.9}&\marksecond{1.1/4.2/17.7}\\
			\midrule
			fisheye3     &3.8/9.5/29.8&1.0/3.6/13.8 &  4.0/10.5/31.6&1.3/4.6/16.4 &  3.4/9.1/28.4&0.8/3.8/13.8& 2.5/6.3/21.9&0.6/2.4/10.1 & 2.8/7.2/22.3&0.9/2.9/12.4& 2.0/5.9/20.0&1.3/4.2/15.0 \\
			+VC1     & \marksecond{5.9/14.7}/39.5&1.5/5.2/17.7 &  \marksecond{6.0}/16.2/43.5&2.0/6.8/21.9 &  \marksecond{5.8/14.7}/39.1&1.8/5.5/18.3&  \marksecond{4.4/10.2}/30.1&1.1/3.3/12.8 & 4.6/11.6/32.0&1.4/4.1/14.4 &  \marksecond{4.3/10.5}/29.7&1.2/3.8/12.3 \\
			+VC2     & 5.2/13.9/\marksecond{41.8}&\marksecond{2.1/6.5/22.5} &  5.9/\marksecond{16.5/46.3}&\marksecond{2.5/8.6/29.1} &  5.4/14.2/\marksecond{40.5}&\marksecond{2.1/7.3/25.9}&  4.3/9.8/\marksecond{34.6}&\marksecond{1.7/5.2/19.5} &  \marksecond{4.7/12.6 /36.8}&\marksecond{2.2/7.1/23.8} & 3.8/\marksecond{10.5 /32.5}&\marksecond{1.6/5.1/20.7}\\
			\midrule
			360 & \markfirst{17.1 / 30.8 / 66.1}&\markfirst{8.5 / 20.1 / 47.5}  & 18.2 /\markfirst{34.6 / 64.2}&\markfirst{7.0 / 18.7 / 45.3} &  15.8 / 31.2 / \markfirst{60.4}&\markfirst{7.0 / 17.8 / 42.8}& \markfirst{17.6 / 31.8 / 68.1}&\markfirst{8.7 / 22.0 / 56.0} &  \markfirst{18.7 / 34.9 / 68.1}&\markfirst{7.3 / 20.0 / 53.4} & \markfirst{ 16.6 / 32.6 / 65.7}&\markfirst{7.1 / 18.7 / 50.4} \\
			\bottomrule
		\end{tabular}
	}
	\caption{Local matching localization results. The average percentage of predictions with high (0.25m, $2^{\circ}$), medium (0.5m, $5^{\circ}$), and low (5m, $10^{\circ}$) accuracy~\cite{sattler2018benchmarking} (higher is better) over four scenes. \marksecond{\footnotesize  $\#$} indicates the highest value for each device w and w/o virtual cameras (VC1, VC2) of each accuracy level. The best results for all devices of each accruacy level are in bold with \markfirst{\footnotesize  $\#$}.}
	\label{tab:res_lm}
\end{table*}

\section{Evaluation}
We provide detailed results for each scene in the dataset and more settings in supplementary material.
\label{sec:exp}
\subsection{Image Retrieval}
We evaluate global descriptors computed by NetVLAD~\cite{arandjelovic2016netvlad}, CosPlace~\cite{Berton_CVPR_2022_CosPlace}, OpenIBL~\cite{ge2020self} and AP-GeM~\cite{GARL17}. The query image is deemed correctly localized if at least one of the top $k$ retrieved database images is within $d = 5 m$ from
the ground truth position of the query for Concourse and $d = 10 m$ for the other three scenes. The image retrieval results are shown in Table~\ref{tab:bs_ir}. Among all global feature descriptor methods, the 360$^\circ$ query exhibits the best precision and recall in most cases, while the pinhole query performs the worst. The remap method (VC1) provides limited improvement for pinhole queries but yields higher improvement for fisheye1, fisheye2, and fisheye3 queries. The reason is that the FoV of pinhole cameras is only 85$^\circ$. Consequently, VC1 results in significant black borders when converting to a 360$^\circ$ image due to the limited coverage.

The rectify method (VC2) significantly improves pinhole, fisheye1, fisheye2, and fisheye3 queries by eliminating the domain gap in IR. However, the pinhole, fisheye1, and fisheye2 queries' recall and precision are still much lower than those of the 360$^\circ$ query. Only the query from fisheye3 (widest FoV)  approaches the performance of 360$^\circ$ query. The domain gap mainly affects the precision and recall of fisheye3. Both remap (VC1) and crop (VC2) significantly improve IR performance for fisheye3. On the other hand, pinhole queries are more prone to being mistaken as erroneous locations with similar structures due to their narrower FoV even there is no cross-device domain gap during IR by applying VC2 (Some figures in supplementary material).

\subsection{Visual Localization}
We compare our approach with the following baselines in two categories: 1) Local feature matching pipelines tailored from HLoc~\cite{sarlin2019coarse}, using different keypoint descriptors (Superpoint (SP)~\cite{detone2018superpoint} and DISK~\cite{tyszkiewicz2020disk}), and matchers (SuperGlue (SG)~\cite{sarlin2020superglue}, follow-up SOTA LightGlue (LG)~\cite{lindenberger2023lightglue}). 2) The
end-to-end APRs: PN~\cite{kendall2015posenet,kendall2017geometric} and MS-T~\cite{shavit2021learning}.

\noindent\textbf{Local feature matching:} During local feature matching, all 360$^\circ$ images are resized to $1228\times614$ because of the tradeoff of time and computation. We report the average results over four scenes in Table~\ref{tab:res_lm}. The 360$^\circ$ query achieves the best performance in three accuracy levels in most cases across all IR, keypoint descriptors, and matchers settings. It is especially more robust in challenging nighttime conditions. VC1 and VC2 techniques improve the recall and precision of IR, increasing the accuracy of 2D-2D matching for all cameras. 
In most cases, the performance at the low accuracy level ($5m,10^\circ$) is correlated with the FoV, where a larger FoV results in higher performance. However, the pinhole query with VC2 during IR performs comparably to the 360$^\circ$ queries at the high ($0.25m,2^\circ$) and median ($0.5m,5^\circ$) accuracy levels. In contrast, query frames from $c_1,c_2$ and $c_3$ demonstrate relatively lower performance at the high and medium accuracy levels.

As observed in Table~\ref{tab:bs_ir}, different IR methods display different performances depending on the type of camera. We thus consider both NetVLAD and CosPlace in visual localization.
In most cases, 360$^\circ$ query frames achieve higher accuracy with CosPlace while pinhole and fisheye query frames have lower accuracy than NetVLAD as shown in Table~\ref{tab:res_lm}.
These results match the precision and recall difference noted in Table~\ref{tab:bs_ir}. We believe that the FoV not only affects the robustness of IR but also has an impact on local 2D-2D matching performance. Pinhole queries suffer from erroneous matches due to interference from symmetrical and repetitive structures, while the larger FoV of fisheye 
 and 360$^\circ$ query frames capture more unique visual features. We provide examples in the supplementary material.

\begin{figure}[!t]
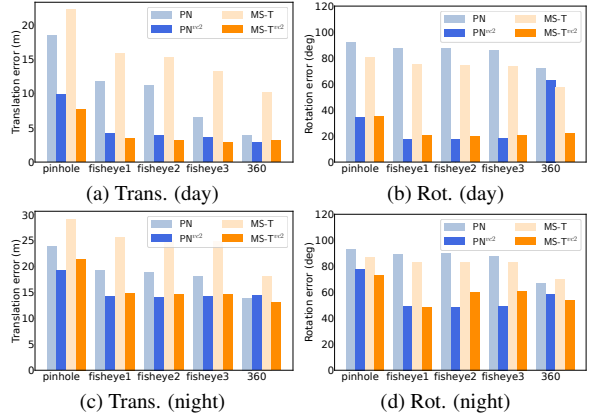

    \centering
    \subfloat[Trans. (day)]{\includegraphics[width=0.45\linewidth]{figure/day_trans.pdf}}\,
    \subfloat[Rot. (day)]{\includegraphics[width=0.45\linewidth]{figure/day_rot.pdf}}\\
    \subfloat[Trans. (night)]
    {\includegraphics[width=0.45\linewidth]{figure/night_trans.pdf}}\,
    \subfloat[Rot. (night)]{\includegraphics[width=0.45\linewidth]{figure/night_rot.pdf}}
    \caption{The average of median translation/rotation errors in ($m/^\circ$) over 4 scenes.}
    \label{fig:bs_apr}
\end{figure}

\noindent\textbf{APR:}
APRs cannot extrapolate well beyond the training set~\cite{sattler2019understanding,ng2021reassessing}.  
cross-device queries further complicate this challenge by introducing an additional dimension of FoV.
Due to the high efficiency of 360$^\circ$ mapping, the training set $\textbf{I}^r$ in 360Loc contains only around one-third of the images compared to datasets~\cite{kendall2015posenet}.
Figure~\ref{fig:bs_apr} shows that when PN and MS-T are trained solely on $\textbf{I}^r$ with only $360^\circ$ images, a smaller domain gap between the query and the $360^\circ$ image yields a lower error. However, when we introduce images from virtual cameras for data augmentation, PN$^{vc2}$ and MS-T$^{vc2}$ exhibit significantly reduced translation and rotation errors across all queries, particularly during daytime. MS-T$^{vc2}$ reduces translation error by up to 79\% and rotation error by up to 72\% compared to MS-T. PN$^{vc2}$ displays similar improvement over PN. In most cases, except for PN$^{vc2}$'s rotation error for the $360^\circ$ queries during daytime, both the 360$^\circ$ and fisheye queries exhibit higher accuracy than the pinhole query on PN$^{vc2}$ and MS-T$^{vc2}$. This suggests that a larger FoV still helps improve visual localization accuracy in challenging scenes. Another interesting finding is that even though the augmented training set $\textbf{I}^r\cup \Psi_c(\textbf{I}^r)$, which includes virtual camera images, does not increase the number of 360$^\circ$ images, the error for the 360$^\circ$ query still decreases. This reduction is particularly noticeable in the case of translation errors during daytime. The result fully demonstrates the utility of employing virtual cameras for data augmentation.

\subsection{Analysis}
Cross-device visual positioning presents significant challenges for IR, local matching, and APRs. Our VC1 and VC2 methods demonstrate practical enhancements in the performance of IR and APR for cross-device scenarios. However, it is essential to note that during the local matching process, the accuracy of matches and the recall and precision of IR for query frames from different cameras may not align perfectly. The chosen IR method and its training noticeably affect accuracy for similar cameras. Fisheye cameras exhibit better performance in IR compared to pinhole cameras. However, pinhole cameras outperform fisheye cameras for high accuracy and median accuracy levels in local matching. This is likely due to existing feature extraction and matching models lacking training data on 360$^\circ$ and fisheye cameras, resulting in less accurate matching. We attribute the inferior performance of pinhole query frames at the low accuracy level to IR's insufficient recall and precision.
Additionally, pinhole queries are more susceptible to interference when there are many
 repetitive and symmetrical features in the scene, even when the retrieved reference image is correct (some example figures in the supplementary material). 
By utilizing VC2 to augment IR and APR's training data, we eliminate the cross-device domain gap. We demonstrate that panoramic perspective and a larger FoV can significantly improve the performance of IR and APRs and find that query frames from 360$^\circ$ camera and ultra-wide FoV cameras are less prone to being misidentified as erroneous locations with similar structures. This result suggests the promising potential of fisheye and 360$^\circ$ cameras as viable sensors for localization tasks in indoor environments with low GPS accuracy.
\section{Conclusion}
360Loc is the first dataset and benchmark that explores the challenge of cross-device visual positioning, involving 360$^\circ$ reference frames, and query frames from pinhole, ultra-wide FoV fisheye, and 360$^\circ$ cameras. 
We first identified the absence of datasets with ground truth 6DoF poses for 360$^\circ$ images, and the limited research on cross-device localization and the robustness of different cameras in ambiguous scenes. 
To address these limitations, we build a dataset with 360$^\circ$ images as reference and query frames from pinhole, ultra-wide FoV fisheye camera and 360$^\circ$ cameras via a \textit{virtual camera} solution. 
This method enables fair comparisons in cross-device visual localization tasks and helps reduce the domain gap between different cameras. 
By evaluating feature-matching-based and pose regression-based methods, we demonstrate the effectiveness of our virtual camera approach and the increased robustness of 360$^\circ$ cameras in visual localization for challenging and ambiguous scenes. 
{
    \small
    \bibliographystyle{ieeenat_fullname}
    \bibliography{main}
}
\clearpage
\setcounter{page}{1}
\maketitlesupplementary
\section{Appendix}
\subsection{Pinhole SfM and 360$^\circ$ Mapping}
\begin{figure*}[!t]
    \centering
    \subfloat[Random pinhole mapping from $c_4$.]{\label{fig:rec_random}\includegraphics[width=0.33\linewidth,height=6cm]{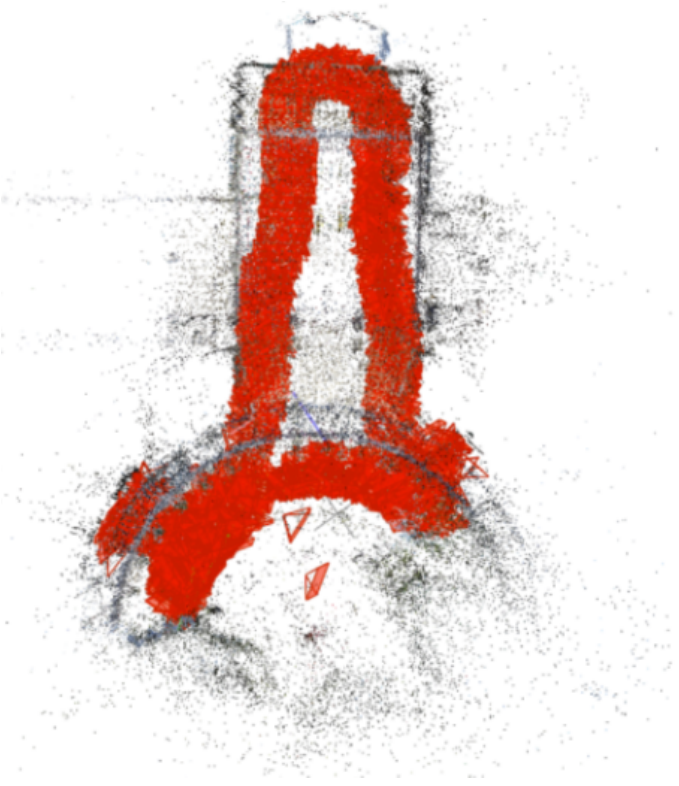}}
    \subfloat[Cube map pinhole mapping.]{\label{fig:rec_cube}\includegraphics[width=0.33\linewidth,height=6cm]{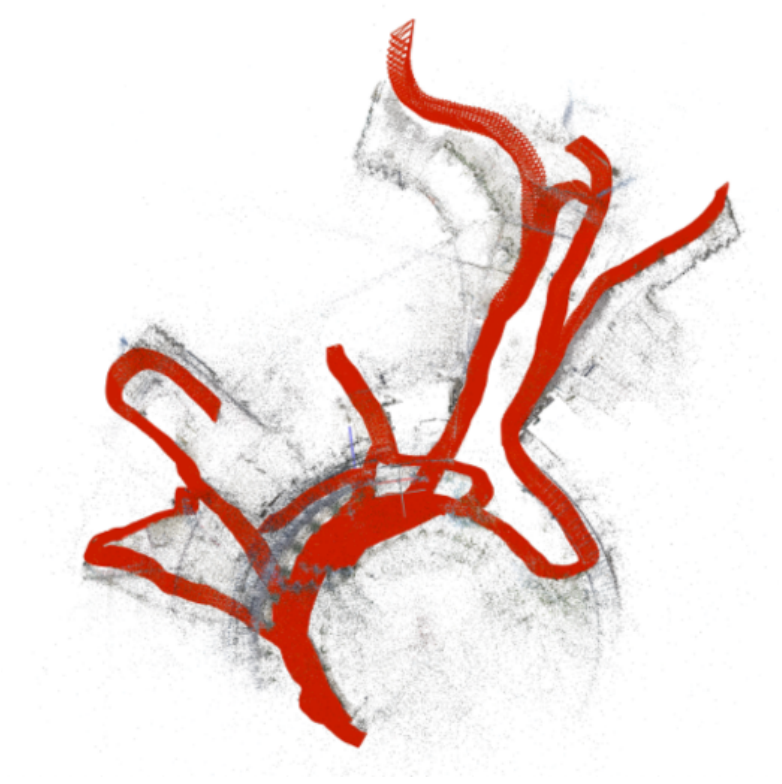}}
    \subfloat[360$^\circ$ mapping.]
    {\label{fig:rec_ours}\includegraphics[width=0.33\linewidth,height=6cm]{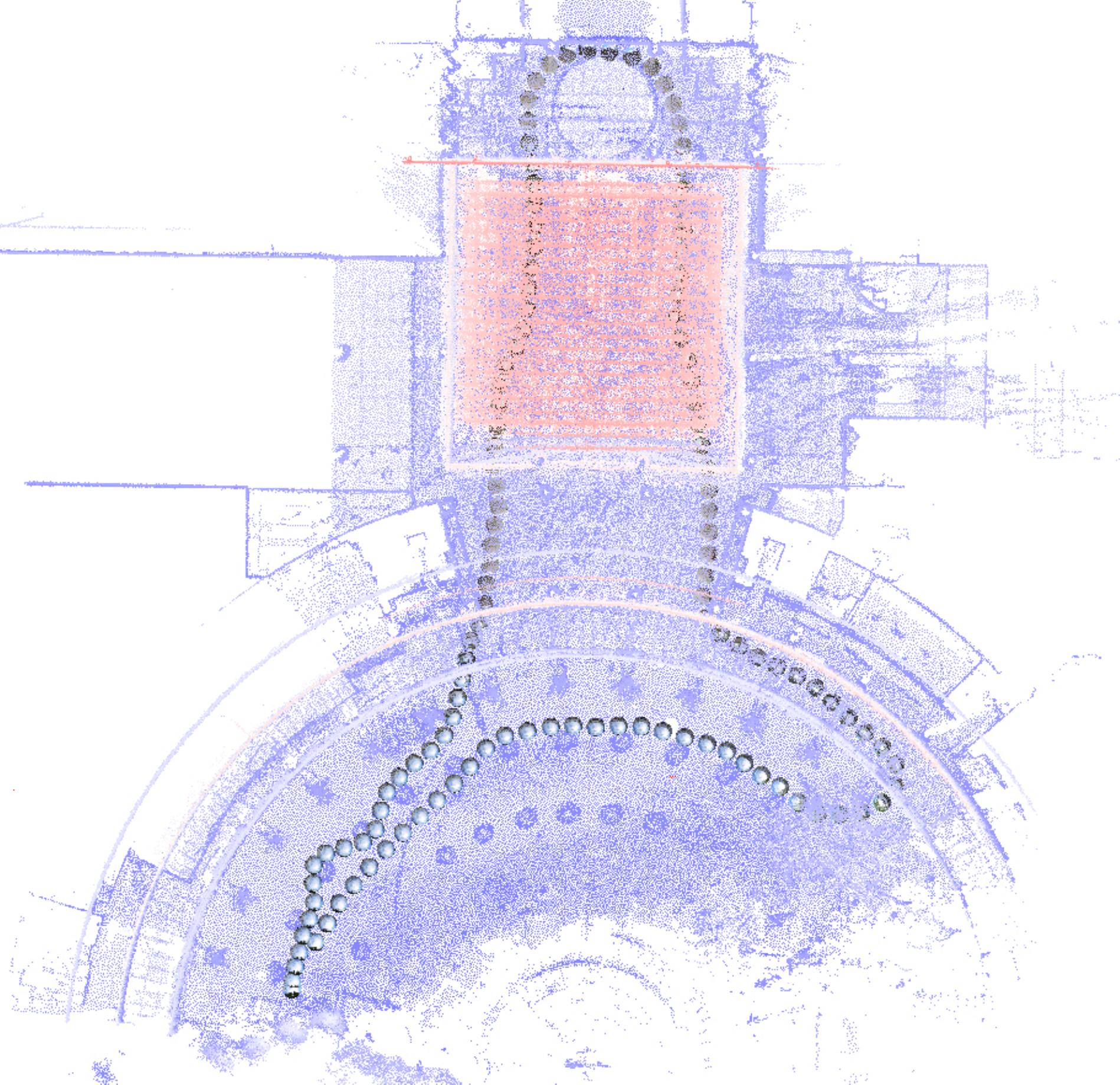}}
    \caption{Subfigure (a)(b) show that the pinhole cameras are incapable of mapping in ambiguous scenes. Ground truth traces are white spheres in Subfigure (c).}
    \label{fig:mapping}
\end{figure*}

Owing to the incompatibility of current SfM methods, 360$^\circ$ images are usually cropped into pinhole images for reconstruction. However, SfM based on pinhole images is vulnerable in particular for 
scenes with symmetric or repetitive features.
To clarify the weakness of Pinhole SfM, we obtain $632\times2$ random cropped images from $c_4$ (see Table~2 in the main paper) and  $632\times5$ cube map images from 632 360$^\circ$ images. The bottom faces of the cube map are discarded because of capturing the ground and platform itself.
After that we use HLoc~\cite{sarlin2019coarse} (NetVLAD~\cite{arandjelovic2016netvlad} top$k$=20, SuperPoint~\cite{detone2018superpoint} + SuperGlue~\cite{sarlin2020superglue}) and COLMAP~\cite{schonberger2016structure} for constructing sparse point cloud models. 
The reconstruction results are demonstrated in Figure~\ref{fig:mapping}. Compared to ground truth Figure~\ref{fig:rec_ours}, using cube map pinhole images Figure~\ref{fig:rec_cube} encounters disaster in trajectory recovery, while the pose estimations in Figure~\ref{fig:rec_random} have large drift.
This example demonstrates that making use of omnidirectional FoV of 360$^\circ$ images is important for accurate and efficient mapping.

\subsection{The Graph-based SLAM for Reference Map Collection}
\label{subsec:mapcollect}
In the process of reference map collection, we utilized a graph-based SLAM technique~\cite{koide2019portable} with a loop detection algorithm. The core of the graph-based method is to minimize the errors between the parameters and the constraints through optimization. We choose the ground plane feature coefficients $\bm{\zeta}_0 = \left[n_x,n_y,n_z,\delta \right]^T=\left[0,0,0,1\right]$ as a constraint to optimize the lidar pose $\bm{\eta}_t = \left[\mathbf{R}_t|\mathbf{t}_t\right]\in SO\left(3\right)\times \mathbb{R}^3$ at frame $t$, ensuring that the ground plane detected in each observation becomes consistent. $\bm{n} = \begin{bmatrix}
n_{x},\ n_{y},\ n_{z}
\end{bmatrix}$is the normal vector of the constrained plane, which points to the $Z$ direction of the $\mathbf{O_w\text{-}XYZ}$, and $\delta$ is the length of the intercept. So, the transformed ground plane constraints in $\mathbf{O_l\text{-}XYZ}$ is:
\begin{equation}
\begin{bmatrix}
n_{x}^{\prime},\ n_{y}^{\prime},\ n_{z}^{\prime}
\end{bmatrix} =
\mathbf{R}_t \cdot \begin{bmatrix}
n_{x},\ n_{y},\ n_{z}
\end{bmatrix},
\label{eq:grdcons1}
\end{equation}
\begin{equation}
    \delta^{\prime} = \delta-\mathbf{t}_t \cdot \begin{bmatrix}
n_{x}^{\prime},\ n_{y}^{\prime},\ n_{z}^{\prime}
\end{bmatrix}^T,
\label{eq:grdcons2}
\end{equation}
where, $\bm{\zeta}_0^{\prime} = \begin{bmatrix}
n_{x}^{\prime},\ n_{y}^{\prime},\ n_{z}^{\prime},\ \delta^{\prime}
\end{bmatrix}$. Thus, the error $\mathbf{\epsilon}_t$ between a lidar pose and the ground plane feature is defined as follows:
\begin{equation}
    \bm{\tau}(\bm{\zeta}) = \begin{bmatrix}
\arctan\left(\frac{n_y}{n_x}\right),\ \arctan\left(\frac{n_z}{|\mathbf{n}|}\right),\ \delta
\end{bmatrix},
\label{eq:grderr1}
\end{equation}
\begin{equation}
    \bm{\epsilon}_t = \bm{\tau}\left(\bm{\zeta}^{\prime}_0\right) - \bm{\tau}\left(\bm{\zeta}_t\right), 
\label{eq:grderr2}
\end{equation}
where $\bm{\zeta}_t$ is the detected ground plane at frame $t$. With this constraint, we can now get the optimized frame-by-frame values of $\bm{\eta}_t$ in real time while collecting.

\subsection{Details about Dataset Collection}
Firstly, it should be noted that during the collection of the reference map, due to the diverse data types and large data volume, we employed a multi-threaded approach in the programming. However, when saving the data, we recorded the timestamps corresponding to their collection triggers and performed offline synchronization.
Here, we provide detailed explanations of several methods used for data collection, as introduced in Section 3 of the main paper.

\begin{figure}
    \centering
    \includegraphics[width=0.9\linewidth]{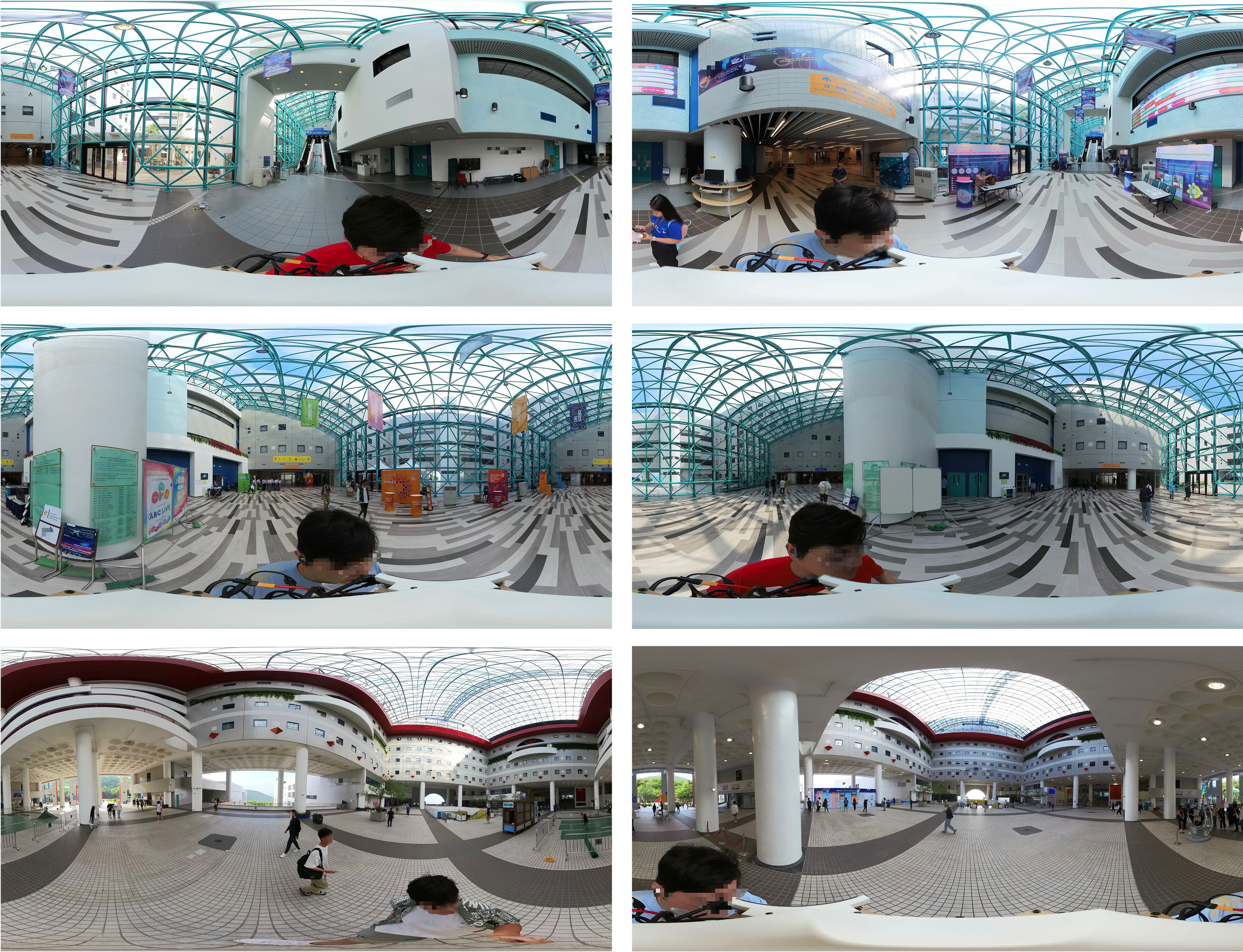}
    \caption{Our datasets feature with appearance changes.}
    \vspace{-0.1cm}
    \label{fig:changes}
\end{figure}

For the image collection process, we set the camera's frame rate to 25 frames per second (fps) and traversed the entire current scene using different routes while manually holding the collection device. This process extended over multiple days and encompassed diverse time periods. Thus, each scene in the dataset includes images in daytime and nighttime. Besides, the images have appearance changes due to the differences in dates as shown in Figure~\ref{fig:changes}.  The data is collected using a handheld device and three participants who recorded sequences were asked to freely walk through each scene with different routines as shown in Figure~\ref{fig:traj}. Collected images feature diverse capture angles, and some may exhibit motion blur. These characteristics make the 360Loc dataset rich, comprehensive, and challenging in nature.

\subsection{Image Retrieval Evaluation}
We evaluate global descriptors computed by NetVLAD~\cite{arandjelovic2016netvlad}, CosPlace~\cite{Berton_CVPR_2022_CosPlace}, OpenIBL~\cite{ge2020self} and AP-GeM~\cite{GARL17}. The query image is deemed correctly localized if at least one of the top $k$ retrieved database images is within $d = 5 m$ from
the ground truth position of the query for Concourse and $d = 10 m$ for the other three scenes. The average IR results over 4 scenes are shown in Table~4 of the main paper.  We provide the IR results of each scene in this supplementary material in Table~\ref{tab:ir_concourse}, Table~\ref{tab:ir_atrium}, Table~\ref{tab:ir_hall}, and Table~\ref{tab:ir_piatrium}. The trend is similar to the Table~4 shown in the main paper.

\begin{figure*}
    \centering
    \includegraphics[width=\linewidth]{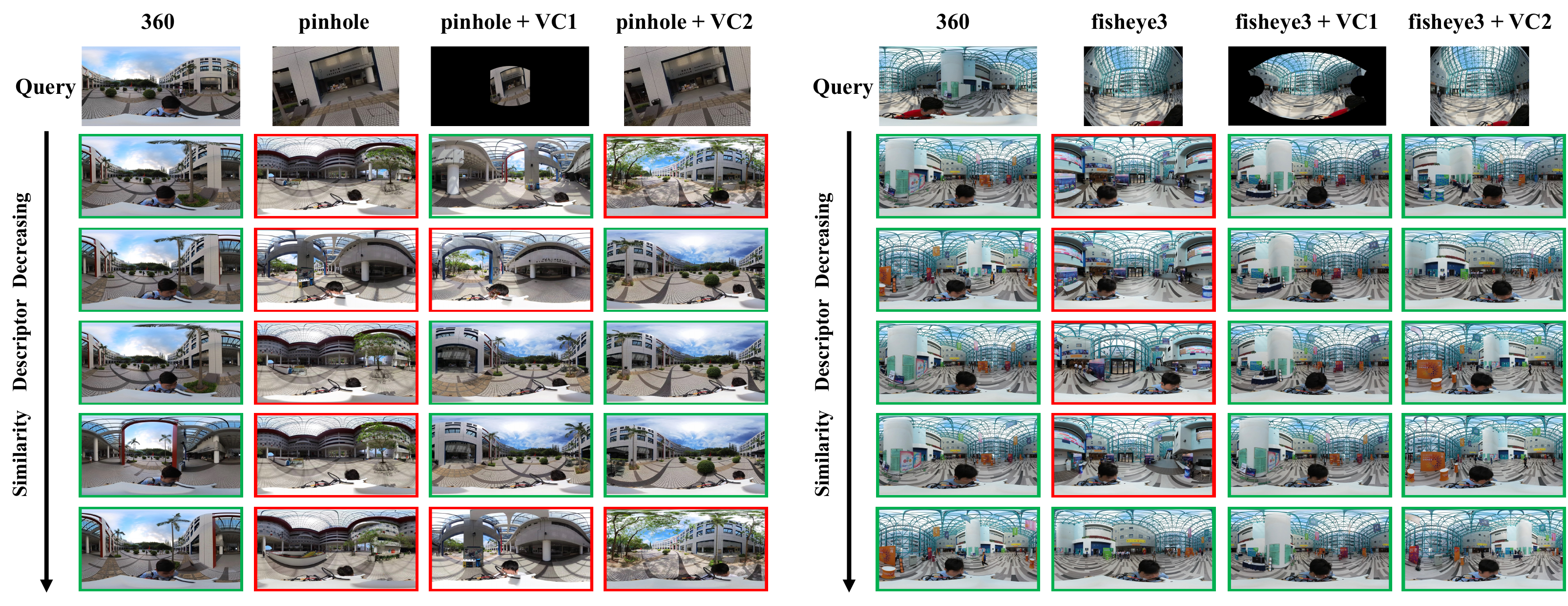}
    \caption{Example queries and top-5 retrieved images. The green boxes represent images that have overlap regions with the query image, and images with red boxes are wrong images.}
    \label{fig:ir_res}
\end{figure*}

Among all global feature descriptor methods, the 360$^\circ$ query exhibits the best precision and recall in most cases, while the pinhole query performs the worst. The performance of recall and precision is correlated with the FoV, where a larger FoV results in higher performance. The remap method (VC1) provides limited improvement for pinhole queries but yields higher improvement for fisheye1, fisheye2, and fisheye3 queries. 
The reason is that the FoV of pinhole cameras is only 85$^\circ$. Consequently, VC1 results in significant black borders when converting to a 360$^\circ$ image due to the limited coverage of the pinhole camera as shown in Figure~\ref{fig:ir_res}.

The rectify method (VC2) significantly improves pinhole, fisheye1, fisheye2, and fisheye3 queries by eliminating the domain gap in IR. However, the pinhole, fisheye1, and fisheye2 queries' recall and precision are still much lower than those of the 360$^\circ$ query. Only the query from fisheye3 (widest FoV)  approaches the performance of 360$^\circ$ query. The domain gap mainly affects the precision and recall of fisheye3. Both remap (VC1) and crop (VC2) significantly improve IR performance for fisheye3. On the other hand, pinhole queries are more prone to being mistaken as erroneous locations with similar structures due to their narrower FoV even there is no cross-device domain gap during IR by applying VC2 as shown in Figure~\ref{fig:ir_res}.

\subsection{Local Feature Matching Evaluation}
The average local feature matching results over 4 scenes are shown in Table~5 of the main paper.  We provide the results of each scene in this supplementary material in Table~\ref{tab:bs_lm_concourse}, Table~\ref{tab:bs_lm_atrium}, Table~\ref{tab:bs_lm_hall}, and Table~\ref{tab:bs_lm_piatrium} with extra SIFT~\cite{lowe2004distinctive} + Nearest Neighbor (NN) setting. The trend is similar to the Table~5 shown in the main paper.

\begin{figure*}
    \centering
    \includegraphics[width=\linewidth]{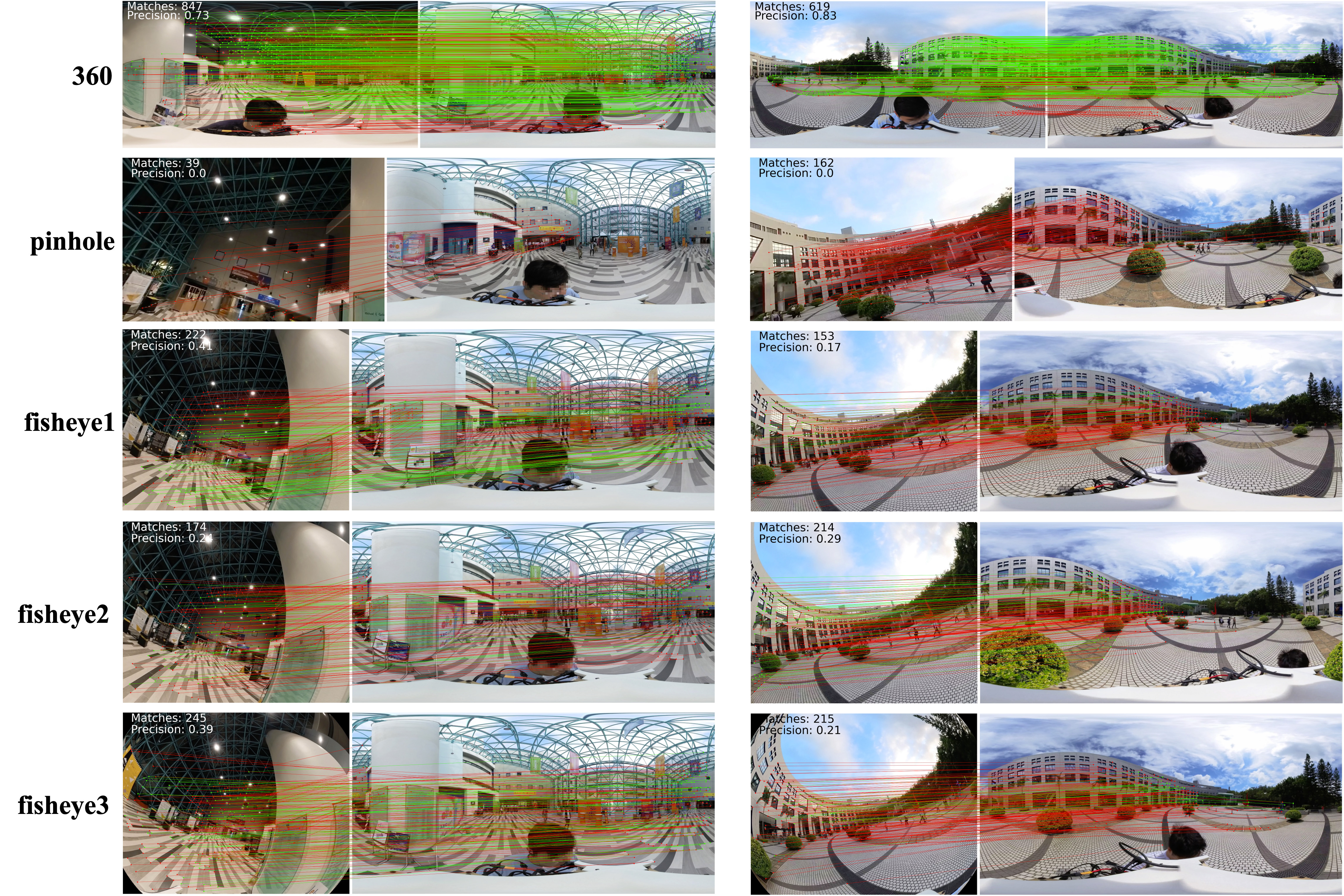}
    \caption{Two groups of local 2D-2D matching pairs (Superpoint + LightGlue~\cite{lindenberger2023lightglue}). For each pair, left side is the query frame and right side is the top-1 retrieved reference 360$^\circ$ image. In this figure, all retrieved images are correct. The reprojection error threshold is 5px (green means correct matches, red means wrong matches).}
    \label{fig:lm}
\end{figure*}

\begin{figure*}
    \centering
    \subfloat[Concourse]{\includegraphics[width=0.23\linewidth,height=2.5cm]{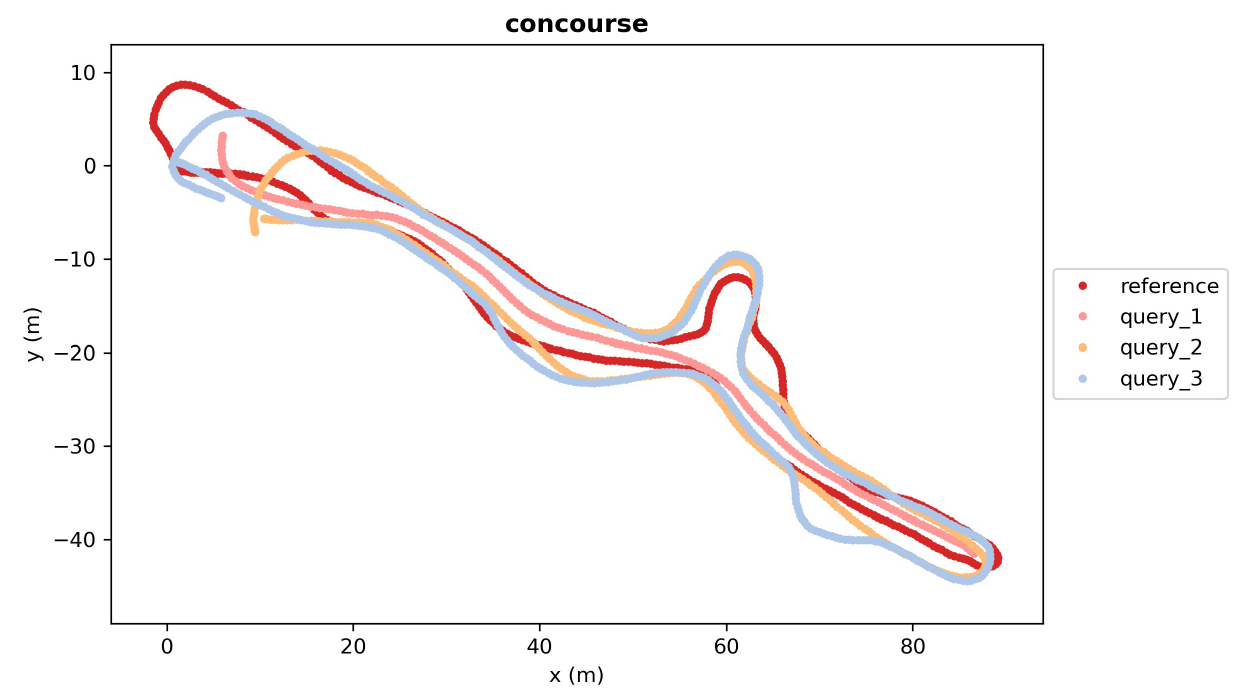}}
    \subfloat[Hall]{\includegraphics[width=0.28\linewidth,height=2.5cm]{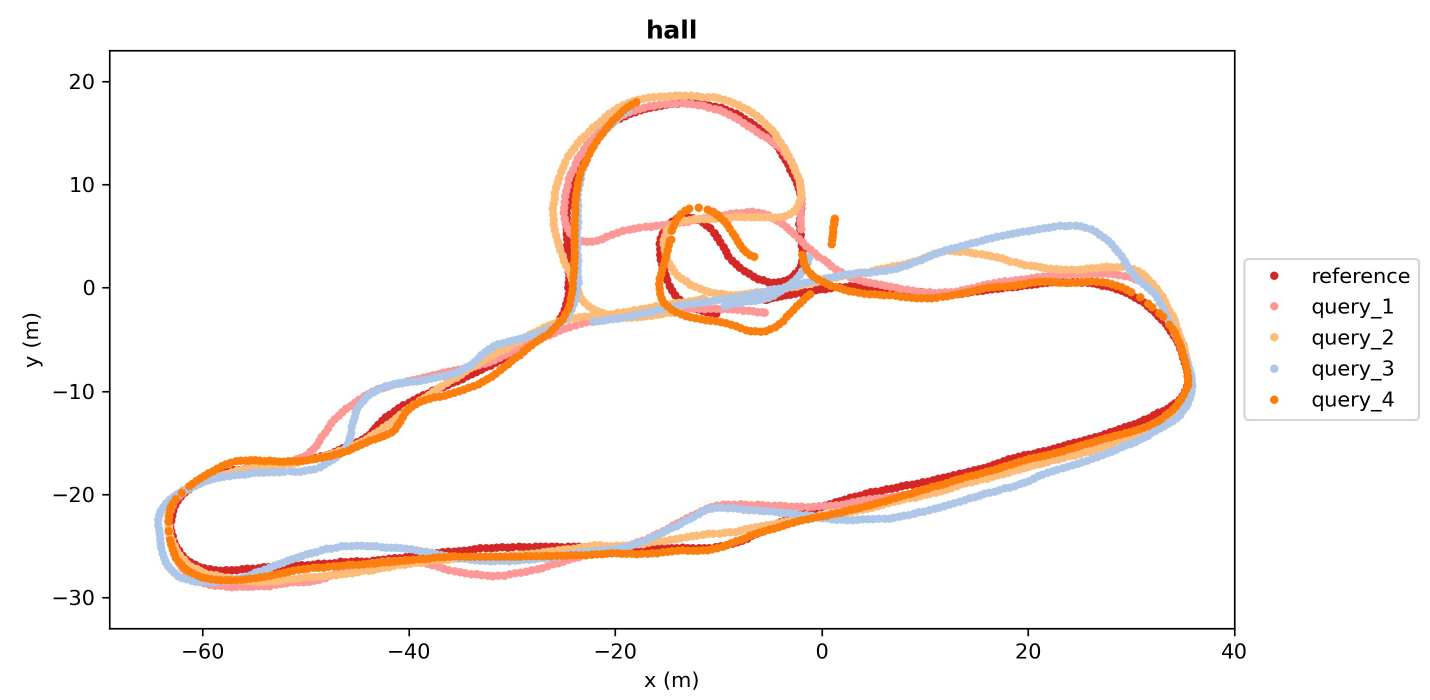}}
    \subfloat[Atrium]{\includegraphics[width=0.17\linewidth,height=2.5cm]{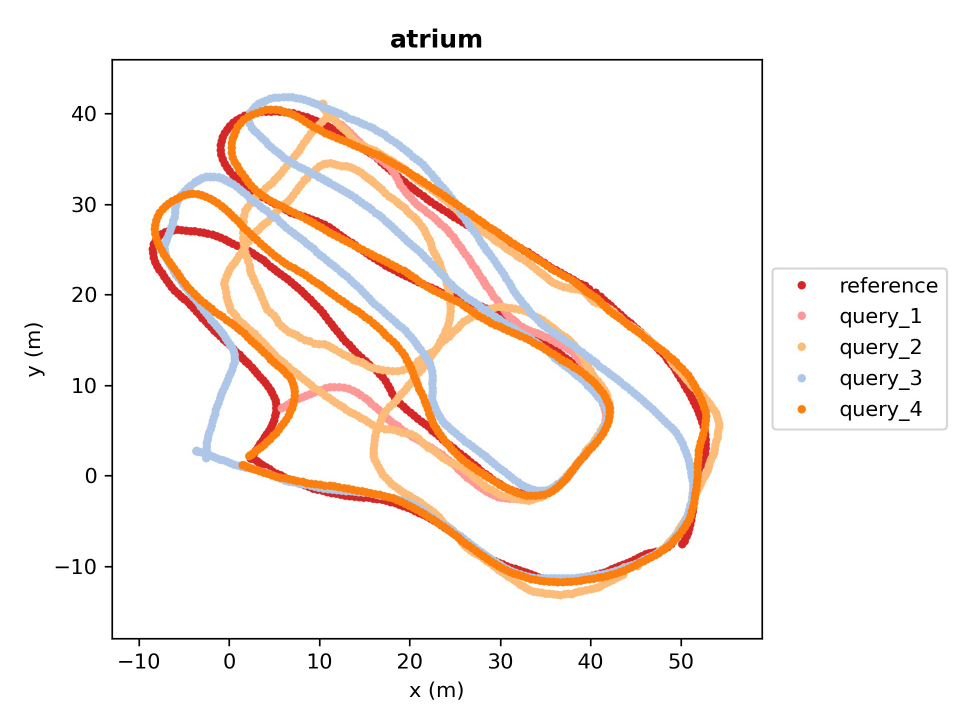}}
    \subfloat[Piatrium]
    {\includegraphics[width=0.28\linewidth,height=2.5cm]{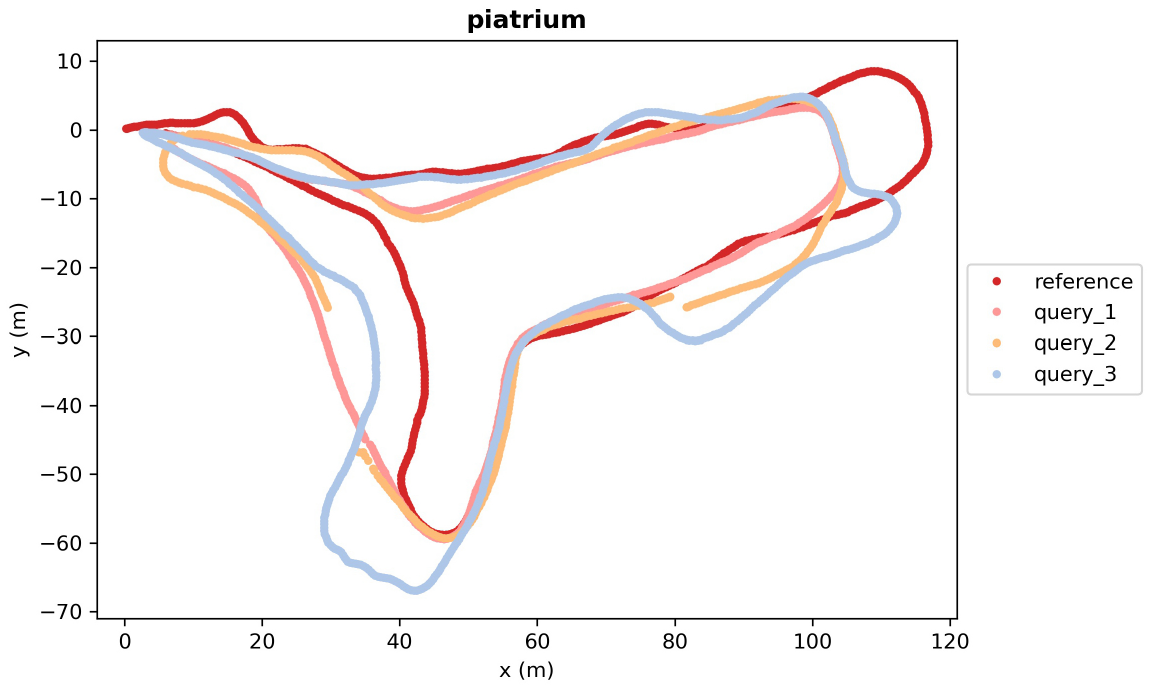}}
    \caption{Trajectories of reference and query.}
    \label{fig:traj}
\end{figure*}

We present two groups of examples of 2D-2D matching. In each group in Figure~\ref{fig:lm}. Query frames from $c_1,c_2,c_3,c_4$ are cropped from the same  360$^\circ$ query, and all corresponding retrieved images overlap with query frames. However, the pinhole query from $c_4$ 
suffers from erroneous matches due to interference from symmetrical and repetitive structures, while fisheye query frames from $c_1,c_2,c_3$, and 360$^\circ$ query frames have better matches as the wide FoV allows to capture more unique features. This finding suggests that cross-device visual localization on a 360-camera database in challenging ambiguous scenarios requires more robust local matching approaches.

\begin{table}[]
    \centering
    \setlength{\tabcolsep}{5pt} 
    \resizebox{\columnwidth}{!}{
        \begin{tabular}{ll|ccccccc}
            \hline
             &  Query & R@1 & R@5 &P@5& R@10& P@10& R@20 &P@20 \\
            \hline
            \multirow{4}{*}{NetVLAD~\cite{arandjelovic2016netvlad}}  
            &  pinhole &0.19 & 0.43 & 0.20 & 0.56 & 0.19 & 0.71 & 0.17  \\
            &  +VC1 & 0.28 & 0.54 & 0.28 & 0.68 & 0.27 & 0.79 & 0.24 \\
            &  +VC2 &\marksecond{0.55} & \marksecond{0.74} & \marksecond{0.52} & \marksecond{0.82} & \marksecond{0.49 }& \marksecond{0.88} & \marksecond{0.41}\\
            &  fisheye1&0.42 & 0.68 & 0.40 & 0.79 & 0.38 & 0.89 & 0.34 \\
            &  +VC1 & 0.53 & 0.77 & 0.50 & 0.86 & 0.47 & 0.92 & 0.41\\
            &  +VC2 &\marksecond{0.79} & \marksecond{0.95} & \marksecond{0.66} & \marksecond{0.97} & \marksecond{0.56} & \marksecond{0.98} & \marksecond{0.46} \\
            &  fisheye2&0.46 & 0.73 & 0.44 & 0.83 & 0.42 & 0.91 & 0.37  \\
            &  +VC1 & 0.54 & 0.78 & 0.52 & 0.87 & 0.49 & 0.93 & 0.43 \\
            &  +VC2 &\marksecond{0.81} & \marksecond{0.95} & \marksecond{0.68} & \marksecond{0.97} & \marksecond{0.58} & \marksecond{0.98} & \marksecond{0.48}\\
            &  fisheye3&0.59 & 0.82 & 0.56 & 0.89 & 0.52 & 0.95 & 0.46 \\
            &  +VC1 &0.65 & 0.86 & 0.62 & 0.92 & 0.58 & 0.96 & \marksecond{0.50} \\
            &  +VC2 &\markfirst{0.82} & \markfirst{0.96} & \markfirst{0.70} & \markfirst{0.98} & \markfirst{0.61} & \markfirst{0.99} & \marksecond{0.50}  \\
            &  360 & 0.68 & 0.71 & 0.65 & 0.71 & \markfirst{0.61} & 0.72 & \markfirst{0.52}\\
            \hline
            \multirow{4}{*}{CosPlace~\cite{Berton_CVPR_2022_CosPlace}}
            &  pinhole &0.16 & 0.29 & 0.15 & 0.38 & 0.15 & 0.50 & 0.15 \\
            &  +VC1 &0.21 & 0.35 & 0.21 & 0.45 & 0.21 & 0.58 & 0.20 \\
            &  +VC2 &\marksecond{0.31} & \marksecond{0.44} & \marksecond{0.31} & \marksecond{0.51} & \marksecond{0.30} & \marksecond{0.60} & \marksecond{0.27} \\
            &  fisheye1& 0.32 & 0.52 & 0.32 & 0.63 & 0.31 & 0.74 & 0.29 \\
            &  +VC1 &0.39 & 0.54 & 0.38 & 0.64 & 0.36 & 0.76 & 0.33  \\
            &  +VC2 & \marksecond{0.62} & \marksecond{0.86} & \marksecond{0.47} & \marksecond{0.94} & \marksecond{0.39} & \marksecond{0.97} & \marksecond{0.31} \\
            &  fisheye2&0.35 & 0.56 & 0.36 & 0.67 & 0.35 & 0.77 & 0.32  \\
            &  +VC1 & 0.39 & 0.54 & 0.38 & 0.64 & 0.37 & 0.76 & 0.34 \\
            &  +VC2 &\marksecond{0.63} & \marksecond{0.88} & \marksecond{0.49} & \marksecond{0.94} & \marksecond{0.41} & \marksecond{0.97} & \marksecond{0.33}\\
            &  fisheye3&0.45 & 0.67 & 0.45 & 0.77 & 0.44 & 0.86 & 0.41 \\
            &  +VC1 &0.52 & 0.67 & 0.50 & 0.76 & 0.49 & 0.86 & 0.44 \\
            &  +VC2 & 0.70 & 0.91 & 0.56 & 0.95 & 0.48 & 0.97 & 0.39 \\
            &  360 &\markfirst{0.94} & \markfirst{0.97} & \markfirst{0.93} & \markfirst{0.98} & \markfirst{0.90} & \markfirst{0.98} & \markfirst{0.78}  \\
            \hline
            \multirow{4}{*}{OpenIBL~\cite{ge2020self}}
            &  pinhole &0.15 & 0.32 & 0.15 & 0.44 & 0.14 & 0.60 & 0.14 \\
            &  +VC1 &0.21 & 0.42 & 0.21 & 0.55 & 0.20 & 0.69 & 0.19 \\
            &  +VC2 &\marksecond{0.55} & \marksecond{0.72} & \marksecond{0.53} & \marksecond{0.78} & \marksecond{0.49} & \marksecond{0.86} & \marksecond{0.42} \\
            &  fisheye1&0.37 & 0.59 & 0.35 & 0.72 & 0.34 & 0.83 & 0.31 \\
            &  +VC1 &0.55 & 0.75 & 0.52 & 0.84 & 0.48 & 0.89 & 0.41 \\
            &  +VC2 &\marksecond{0.75} & \marksecond{0.92} & \marksecond{0.62} & \marksecond{0.96} & \marksecond{0.53} & \marksecond{0.98} & \marksecond{0.43} \\
            &  fisheye2&0.41 & 0.66 & 0.39 & 0.77 & 0.37 & 0.86 & 0.34 \\
            &  +VC1 &0.58 & 0.78 & 0.55 & 0.85 & 0.51 & 0.90 & 0.44 \\
            &  +VC2 &\marksecond{0.77} & \marksecond{0.93} & \marksecond{0.63} & \marksecond{0.97} & \marksecond{0.55} & \marksecond{0.99} & \marksecond{0.45} \\
            &  fisheye3&0.54 & 0.78 & 0.52 & 0.87 & 0.49 & 0.93 & 0.44 \\
            &  +VC1 &0.69 & 0.87 & 0.67 & 0.92 & \marksecond{0.62} & 0.96 & \marksecond{0.53}\\
            &  +VC2 &\marksecond{0.80} & \marksecond{0.94} & \marksecond{0.66} & \marksecond{0.97} & 0.57 & \markfirst{0.99} & 0.47 \\
            &  360 & \markfirst{0.91} & \markfirst{0.96} & \markfirst{0.90} & \markfirst{0.97} & \markfirst{0.84} & 0.98 & \markfirst{0.72} \\
            \hline
            \multirow{4}{*}{AP-GeM~\cite{GARL17}}
            &  pinhole &0.17 & 0.36 & 0.18 & 0.47 & 0.18 & 0.61 & 0.17 \\
            &  +VC1 & 0.26 & 0.46 & 0.26 & 0.58 & 0.25 & 0.72 & 0.23  \\
            &  +VC2 &\marksecond{0.50} & \marksecond{0.71} & \marksecond{0.48} & \marksecond{0.79} & \marksecond{0.45} & \marksecond{0.88} & \marksecond{0.39} \\
            &  fisheye1&0.37 & 0.60 & 0.35 & 0.71 & 0.33 & 0.80 & 0.30 \\
            &  +VC1 &0.46 & 0.68 & 0.45 & 0.78 & 0.42 & 0.86 & \marksecond{0.38} \\
            &  +VC2 &\marksecond{0.66} & \marksecond{0.90} & \marksecond{0.54} & \marksecond{0.95} & \marksecond{0.47} & \marksecond{0.98} & \marksecond{0.38} \\
            &  fisheye2&0.40 & 0.64 & 0.38 & 0.73 & 0.36 & 0.82 & 0.33\\
            &  +VC1 &0.49 & 0.71 & 0.48 & 0.80 & 0.45 & 0.88 & \marksecond{0.40} \\
            &  +VC2 & \marksecond{0.68} & \marksecond{0.91} & \marksecond{0.56} & \marksecond{0.96} & \marksecond{0.49} & \marksecond{0.98} & \marksecond{0.40} \\
            &  fisheye3&0.47 & 0.69 & 0.45 & 0.77 & 0.42 & 0.85 & 0.37 \\
            &  +VC1 &0.59 & 0.78 & 0.56 & 0.85 & \marksecond{0.53} & 0.92 & \marksecond{0.46} \\
            &  +VC2 &\marksecond{0.69} & \marksecond{0.92} & \marksecond{0.58} & \marksecond{0.96} & 0.52 & \marksecond{0.97} & 0.43 \\
            &  360 &\markfirst{0.84} & \markfirst{0.94} & \markfirst{0.80} & \markfirst{0.97} & \markfirst{0.75} & \markfirst{0.98} & \markfirst{0.64} \\
            \hline
        \end{tabular}
    }
    \caption{\textit{Concourse}. Image retrieval results based on 360$^\circ$ reference database for the top $k$ retrieved images,
$k$ = 1, 5, 10, 20. \marksecond{\footnotesize $\#$} indicates the highest value  of R@$k$ and P@$k$ for each device w and w/o virtual cameras (VC1, VC2). Best results for all devices of R@$k$ and P@$k$ are in bold with \markfirst{\footnotesize $\#$}.}
     \label{tab:ir_concourse}
\end{table}

\begin{table}[]
    \centering
    \setlength{\tabcolsep}{5pt} 
    \resizebox{\columnwidth}{!}{
        \begin{tabular}{ll|ccccccc}
            \hline
             &  Query & R@1 & R@5 &P@5& R@10& P@10& R@20 &P@20 \\
            \hline
            \multirow{4}{*}{NetVLAD~\cite{arandjelovic2016netvlad}}  
            &  pinhole &0.30 & 0.55 & 0.29 & 0.69 & 0.28 & 0.81 & 0.25  \\
            &  +VC1 &0.24 & 0.43 & 0.24 & 0.53 & 0.23 & 0.64 & 0.21 \\
            &  +VC2 &\marksecond{0.55} & \marksecond{0.69} & \marksecond{0.53} & \marksecond{0.76} & \marksecond{0.52} & \marksecond{0.84} & \marksecond{0.48}  \\
            &  fisheye1&0.56 & 0.79 & 0.54 & 0.87 & 0.51 & 0.93 & 0.46 \\
            &  +VC1 &0.57 & 0.75 & 0.55 & 0.82 & 0.53 & 0.89 & 0.48\\
            &  +VC2 &\marksecond{0.76} & \marksecond{0.93} & \marksecond{0.70} & \marksecond{0.98} & \marksecond{0.65} & \marksecond{0.99} & \marksecond{0.57} \\
            &  fisheye2&0.60 & 0.82 & 0.57 & 0.89 & 0.55 & 0.94 & 0.50 \\
            &  +VC1 &0.61 & 0.78 & 0.59 & 0.85 & 0.57 & 0.90 & 0.52  \\
            &  +VC2 & \marksecond{0.77} & \marksecond{0.94} & \marksecond{0.72} & \marksecond{0.98} & \marksecond{0.67} & \marksecond{0.99} & \marksecond{0.59}  \\
            &  fisheye3&0.69 & 0.87 & 0.66 & 0.92 & 0.64 & 0.96 & 0.58 \\
            &  +VC1 &0.70 & 0.85 & 0.69 & 0.91 & 0.66 & 0.95 & 0.61\\
            &  +VC2 &\marksecond{0.81} & \markfirst{0.96} & \marksecond{0.75} & \markfirst{0.99} & \marksecond{0.70} & \markfirst{1.00} & \marksecond{0.63} \\
            &  360 &\markfirst{0.90} & \markfirst{0.96} & \markfirst{0.89} & 0.98 & \markfirst{0.86} & \markfirst{1.00} & \markfirst{0.81}  \\
            \hline
            \multirow{4}{*}{CosPlace~\cite{Berton_CVPR_2022_CosPlace}}
            &  pinhole &0.19 & 0.32 & 0.19 & 0.42 & 0.19 & 0.55 & 0.19 \\
            &  +VC1 &0.25 & 0.37 & 0.24 & 0.46 & 0.24 & 0.56 & 0.23 \\
            &  +VC2  &\marksecond{0.35} & \marksecond{0.43} & \marksecond{0.34} & \marksecond{0.49} & \marksecond{0.33} & \marksecond{0.57} & \marksecond{0.31}\\
            &  fisheye1& 0.39 & 0.54 & 0.38 & 0.65 & 0.38 & 0.76 & 0.36 \\
            &  +VC1 &0.38 & 0.52 & 0.37 & 0.60 & 0.36 & 0.68 & 0.33\\
            &  +VC2  & \marksecond{0.66} & \marksecond{0.88} & \marksecond{0.56} & \marksecond{0.94} & \marksecond{0.49} & \marksecond{0.98} & \marksecond{0.41}\\
            &  fisheye2&0.41 & 0.59 & 0.42 & 0.68 & 0.42 & 0.79 & 0.40 \\
            &  +VC1 &0.40 & 0.54 & 0.39 & 0.61 & 0.37 & 0.70 & 0.34 \\
            &  +VC2  &\marksecond{0.67} & \marksecond{0.89} & \marksecond{0.59} & \marksecond{0.95} & \marksecond{0.52} & 
            \marksecond{0.98} & \marksecond{0.43}\\
            &  fisheye3&0.54 & 0.70 & 0.54 & 0.78 & 0.53 & 0.87 & \marksecond{0.50} \\
            &  +VC1 &0.52 & 0.67 & 0.53 & 0.74 & 0.51 & 0.81 & 0.48\\
            &  +VC2  &\marksecond{0.73} & \marksecond{0.91} & \marksecond{0.64} & \marksecond{0.96} & \marksecond{0.58} & 
            \marksecond{0.99} & 0.49 \\
            &  360 &\markfirst{0.96} & \markfirst{0.98} & \markfirst{0.96} & \markfirst{0.99} & \markfirst{0.94} & \markfirst{1.00} & \markfirst{0.89}\\
            \hline
            \multirow{4}{*}{OpenIBL~\cite{ge2020self}}
            &  pinhole &0.23 & 0.45 & 0.23 & 0.58 & 0.23 & 0.71 & 0.21 \\
            &  +VC1 &0.24 & 0.40 & 0.22 & 0.50 & 0.21 & 0.61 & 0.20\\
            &  +VC2 &\marksecond{0.53} & \marksecond{0.69} & \marksecond{0.52} & \marksecond{0.76} & \marksecond{0.50} & \marksecond{0.84} & \marksecond{0.45}  \\
            &  fisheye1&0.48 & 0.71 & 0.47 & 0.81 & 0.45 & 0.89 & 0.41 \\
            &  +VC1 &0.55 & 0.73 & 0.53 & 0.81 & 0.50 & 0.88 & 0.45\\
            &  +VC2 &\marksecond{0.77} & \marksecond{0.94} & \marksecond{0.69} & \marksecond{0.98} & \marksecond{0.62} & \marksecond{0.99} & \marksecond{0.52}\\
             &  fisheye2& 0.53 & 0.75 & 0.52 & 0.84 & 0.49 & 0.91 & 0.44\\
            &  +VC1 &0.59 & 0.76 & 0.57 & 0.84 & 0.54 & 0.90 & 0.48 \\
            &  +VC2 &\marksecond{0.79} & \marksecond{0.95} & \marksecond{0.71} & \marksecond{0.98} & \marksecond{0.64} & \markfirst{1.00} & \marksecond{0.55}\\
            &  fisheye3&0.65 & 0.84 & 0.63 & 0.91 & 0.60 & 0.95 & 0.54\\
            &  +VC1 &0.72 & 0.86 & 0.70 & 0.92 & 0.67 & 0.96 & \marksecond{0.60} \\
            &  +VC2 &\marksecond{0.82} & \marksecond{0.96} & \marksecond{0.74} & \markfirst{0.99} & \marksecond{0.68} & \markfirst{1.00} & 0.59 \\
            &  360 &\markfirst{0.95} & \markfirst{0.98} & \markfirst{0.94} & \markfirst{0.99} & \markfirst{0.91} & \markfirst{1.00} & \markfirst{0.83}\\
            \hline
            \multirow{4}{*}{AP-GeM~\cite{GARL17}}
            &  pinhole &0.31 & 0.50 & 0.30 & 0.61 & 0.29 & 0.72 & 0.27 \\
            &  +VC1 & 0.34 & 0.54 & 0.33 & 0.64 & 0.32 & 0.74 & 0.29 \\
            &  +VC2 &\marksecond{0.63} & \marksecond{0.79} & \marksecond{0.62} & \marksecond{0.87} & \marksecond{0.59} & \marksecond{0.93} & \marksecond{0.54}\\
            &  fisheye1&0.49 & 0.69 & 0.47 & 0.78 & 0.45 & 0.86 & 0.41 \\
            &  +VC1 & 0.52 & 0.70 & 0.50 & 0.78 & 0.47 & 0.85 & 0.43  \\
            &  +VC2 &\marksecond{0.75} & \marksecond{0.94} & \marksecond{0.69} & 
            \markfirst{0.98} & \marksecond{0.63} & \markfirst{0.99} & \marksecond{0.55} \\
            &  fisheye2&0.52 & 0.71 & 0.50 & 0.80 & 0.48 & 0.87 & 0.44 \\
            &  +VC1 &0.55 & 0.73 & 0.53 & 0.81 & 0.50 & 0.87 & 0.46 \\
            &  +VC2 &\marksecond{0.77} & \marksecond{0.94} & \marksecond{0.70} & \markfirst{0.98} & \marksecond{0.64} & \markfirst{0.99} & \marksecond{0.56} \\
            &  fisheye3&0.58 & 0.78 & 0.56 & 0.86 & 0.54 & 0.92 & 0.49 \\
            &  +VC1 &0.66 & 0.83 & 0.64 & 0.89 & 0.61 & 0.93 & 0.56  \\
            &  +VC2 &\marksecond{0.78} & \marksecond{0.95} & \marksecond{0.72} & \markfirst{0.98} & \marksecond{0.67} & \markfirst{0.99} & \marksecond{0.58} \\
            &  360 &\markfirst{0.90} & \markfirst{0.96} & \markfirst{0.87} & \markfirst{0.98} & \markfirst{0.83} & \markfirst{0.99} & \markfirst{0.76} \\
            \hline
        \end{tabular}
    }
    \caption{\textit{Hall}. Image retrieval results based on 360$^\circ$ reference database for the top $k$ retrieved images,
$k$ = 1, 5, 10, 20. \marksecond{\footnotesize $\#$} indicates the highest value  of R@$k$ and P@$k$ for each device w and w/o virtual cameras (VC1, VC2). Best results for all devices of R@$k$ and P@$k$ are in bold with \markfirst{\footnotesize $\#$}.}
\label{tab:ir_hall}
\end{table}

\begin{table}[]
    \centering
    \setlength{\tabcolsep}{5pt} 
    \resizebox{\columnwidth}{!}{
        \begin{tabular}{ll|ccccccc}
            \hline
             &  Query & R@1 & R@5 &P@5& R@10& P@10& R@20 &P@20 \\
            \hline
            \multirow{4}{*}{NetVLAD~\cite{arandjelovic2016netvlad}}  
            &  pinhole &0.22 & 0.44 & 0.21 & 0.57 & 0.21 & 0.69 & 0.20\\
            &  +VC1 &0.24 & 0.44 & 0.24 & 0.57 & 0.23 & 0.72 & 0.22\\
            &  +VC2 &\marksecond{0.48} & \marksecond{0.67} & \marksecond{0.47} & \marksecond{0.76} & \marksecond{0.46} & \marksecond{0.85} & \marksecond{0.42} \\
            &  fisheye1&0.39 & 0.63 & 0.37 & 0.74 & 0.35 & 0.85 & 0.33\\
            &  +VC1 & 0.48 & 0.71 & 0.46 & 0.82 & 0.45 & 0.90 & 0.41 \\
            &  +VC2 &\marksecond{0.75} & \marksecond{0.94} & \marksecond{0.63} & \marksecond{0.98} & \marksecond{0.55} & \marksecond{0.99} & \marksecond{0.47}\\
            &  fisheye2& 0.42 & 0.66 & 0.40 & 0.78 & 0.39 & 0.88 & 0.36\\
            &  +VC1 & 0.51 & 0.74 & 0.50 & 0.84 & 0.48 & 0.91 & 0.43 \\
            &  +VC2 &\marksecond{0.75} & \marksecond{0.95} & \marksecond{0.65} & \marksecond{0.98} & \marksecond{0.58} & \markfirst{1.00} & \marksecond{0.49}\\
            &  fisheye3&0.52 & 0.77 & 0.51 & 0.86 & 0.48 & 0.93 & 0.45\\
            &  +VC1 &0.60 & 0.81 & 0.59 & 0.89 & 0.56 & 0.95 & 0.51 \\
            &  +VC2 &\marksecond{0.79} & \marksecond{0.96} & \marksecond{0.68} & \markfirst{0.99} & \marksecond{0.60} & \markfirst{1.00} & \marksecond{0.52} \\
            &  360 & \markfirst{0.85} & \markfirst{0.95} & \markfirst{0.82} & 0.98 & \markfirst{0.79} & \markfirst{1.00} & \markfirst{0.70} \\
            \hline
            \multirow{4}{*}{CosPlace~\cite{Berton_CVPR_2022_CosPlace}}
            &  pinhole &0.15 & 0.23 & 0.15 & 0.29 & 0.15 & 0.37 & 0.14\\
            &  +VC1 &0.21 & 0.33 & 0.21 & 0.42 & 0.21 & 0.52 & 0.21\\
            &  +VC2 &\marksecond{0.34} & \marksecond{0.43} & \marksecond{0.34} & \marksecond{0.49} & \marksecond{0.33} & \marksecond{0.59} & \marksecond{0.32}\\
            &  fisheye1&0.22 & 0.32 & 0.22 & 0.39 & 0.22 & 0.49 & 0.21\\
            &  +VC1 &0.36 & 0.48 & 0.36 & 0.55 & 0.36 & 0.63 & 0.34 \\
            &  +VC2 &\marksecond{0.65} & \marksecond{0.88} & \marksecond{0.53} & \marksecond{0.94} & \marksecond{0.45} & \marksecond{0.98} & \marksecond{0.36}\\
            &  fisheye2&0.23 & 0.35 & 0.24 & 0.41 & 0.23 & 0.52 & 0.22\\
            &  +VC1 &0.38 & 0.50 & 0.37 & 0.57 & 0.37 & 0.64 & 0.36 \\
            &  +VC2 &\marksecond{0.67} & \marksecond{0.89} & \marksecond{0.55} & \marksecond{0.95} & \marksecond{0.46} & \marksecond{0.98} & \marksecond{0.37}\\
            &  fisheye3& 0.30 & 0.44 & 0.31 & 0.52 & 0.30 & 0.63 & 0.30\\
            &  +VC1 & 0.49 & 0.60 & 0.48 & 0.66 & 0.47 & 0.73 & 0.45 \\
            &  +VC2 & \marksecond{0.72} & \marksecond{0.91} & \marksecond{0.59} & \markfirst{0.96} & \marksecond{0.50} & \markfirst{0.99} & \marksecond{0.41} \\
            &  360 &\markfirst{0.92} & \markfirst{0.94} & \markfirst{0.91} & \markfirst{0.96} & \markfirst{0.90} & 0.97 & \markfirst{0.85}\\
            \hline
            \multirow{4}{*}{OpenIBL~\cite{ge2020self}}
            &  pinhole &0.17 & 0.35 & 0.17 & 0.46 & 0.17 & 0.60 & 0.16 \\
            &  +VC1 & 0.23 & 0.42 & 0.23 & 0.54 & 0.23 & 0.68 & 0.22 \\
            &  +VC2 & \marksecond{0.51} & \marksecond{0.69} & \marksecond{0.50} & \marksecond{0.78} & \marksecond{0.48} & \marksecond{0.86} & \marksecond{0.44}\\
            &  fisheye1&0.33 & 0.53 & 0.31 & 0.64 & 0.30 & 0.76 & 0.28 \\
            &  +VC1 & 0.52 & 0.72 & 0.51 & 0.82 & 0.49 & 0.90 & 0.45\\
            &  +VC2 &\marksecond{0.77} & \marksecond{0.95} & \marksecond{0.64} & \marksecond{0.98} & \marksecond{0.55} & \markfirst{1.00} & \marksecond{0.46} \\
            &  fisheye2&0.35 & 0.57 & 0.34 & 0.69 & 0.33 & 0.80 & 0.31\\
            &  +VC1 &0.56 & 0.74 & 0.54 & 0.83 & 0.52 & 0.90 & 0.48\\
            &  +VC2 &\marksecond{0.80} & \marksecond{0.96} & \marksecond{0.66} & \marksecond{0.98} & \marksecond{0.57} & \markfirst{1.00} & \marksecond{0.47}\\
            &  fisheye3&0.48 & 0.70 & 0.46 & 0.81 & 0.45 & 0.91 & 0.41 \\
            &  +VC1 &0.65 & 0.81 & 0.64 & 0.89 & 0.61 & 0.94 & 0.56\\
            &  +VC2 & \marksecond{0.83} & \markfirst{0.97} & \marksecond{0.71} & \markfirst{0.99} & \marksecond{0.61} & \markfirst{1.00} & \marksecond{0.51} \\
            &  360 &\markfirst{0.90} & \markfirst{0.97} & \markfirst{0.89} & 0.98 & \markfirst{0.85} & \markfirst{1.00} & \markfirst{0.75} \\
            \hline
            \multirow{4}{*}{AP-GeM~\cite{GARL17}}
            &  pinhole &0.15 & 0.29 & 0.15 & 0.38 & 0.15 & 0.49 & 0.15\\
            &  +VC1 & 0.19 & 0.34 & 0.19 & 0.44 & 0.19 & 0.56 & 0.19\\
            &  +VC2 &\marksecond{0.46} & \marksecond{0.64} & \marksecond{0.45} & \marksecond{0.74} & \marksecond{0.44} & \marksecond{0.83} & \marksecond{0.40}\\
            &  fisheye1&0.22 & 0.41 & 0.22 & 0.52 & 0.22 & 0.63 & 0.22 \\
            &  +VC1 & 0.36 & 0.57 & 0.36 & 0.67 & 0.35 & 0.79 & 0.34 \\
            &  +VC2 &\marksecond{0.63} & \marksecond{0.89} & \marksecond{0.55} & \marksecond{0.95} & \marksecond{0.49} & \marksecond{0.98} & \marksecond{0.43}\\
            &  fisheye2&0.25 & 0.44 & 0.24 & 0.55 & 0.24 & 0.67 & 0.24\\
            &  +VC1 &0.40 & 0.60 & 0.39 & 0.70 & 0.38 & 0.81 & 0.36 \\
            &  +VC2 &\marksecond{0.64} & \marksecond{0.90} & \marksecond{0.56} & \marksecond{0.96} & \marksecond{0.50} & \marksecond{0.99} & \marksecond{0.44} \\
            &  fisheye3&0.33 & 0.55 & 0.33 & 0.66 & 0.32 & 0.78 & 0.31\\
            &  +VC1 & 0.48 & 0.68 & 0.48 & 0.78 & 0.46 & 0.88 & 0.44\\
            &  +VC2 &\marksecond{0.65} & \markfirst{0.92} & \marksecond{0.57} & \markfirst{0.97} & \marksecond{0.51} & \markfirst{0.99} & \marksecond{0.45}\\
            &  360 & \markfirst{0.73} &\marksecond{ 0.88} & \markfirst{0.72} & \marksecond{0.94} & \markfirst{0.68} & \marksecond{0.98} & \markfirst{0.62} \\
            \hline
        \end{tabular}
    }
    \caption{\textit{Atrium}. Image retrieval results based on 360$^\circ$ reference database for the top $k$ retrieved images,
$k$ = 1, 5, 10, 20. \marksecond{\footnotesize $\#$} indicates the highest value  of R@$k$ and P@$k$ for each device w and w/o virtual cameras (VC1, VC2). Best results for all devices of R@$k$ and P@$k$ are in bold with \markfirst{\footnotesize $\#$}.}
\label{tab:ir_atrium}
\end{table}

\begin{table}[]
    \centering
    \setlength{\tabcolsep}{5pt} 
    \resizebox{\columnwidth}{!}{
        \begin{tabular}{ll|ccccccc}
            \hline
             &  Query & R@1 & R@5 &P@5& R@10& P@10& R@20 &P@20 \\
            \hline
            \multirow{4}{*}{NetVLAD~\cite{arandjelovic2016netvlad}}  
            &  pinhole &0.20 & 0.39 & 0.19 & 0.51 & 0.18 & 0.64 & 0.17 \\
            &  +VC1 & 0.21 & 0.38 & 0.20 & 0.50 & 0.20 & 0.61 & 0.19 \\
            &  +VC2 & \marksecond{0.43} & \marksecond{0.59} & \marksecond{0.42} & \marksecond{0.67} & \marksecond{0.41} & \marksecond{0.76} & \marksecond{0.38} \\
            &  fisheye1&0.32 & 0.56 & 0.31 & 0.67 & 0.30 & 0.78 & 0.28\\
            &  +VC1 &0.45 & 0.64 & 0.44 & 0.72 & 0.42 & 0.80 & 0.39 \\
            &  +VC2 & \marksecond{0.62} & \marksecond{0.82} & \marksecond{0.54} & \marksecond{0.87} & \marksecond{0.49} & \marksecond{0.90} & \marksecond{0.42} \\
            &  fisheye2&0.34 & 0.59 & 0.34 & 0.71 & 0.33 & 0.81 & 0.31 \\
            &  +VC1 &0.49 & 0.66 & 0.47 & 0.74 & 0.44 & 0.81 & 0.41\\
            &  +VC2 & \marksecond{0.64} & \marksecond{0.84} & \marksecond{0.56} & \marksecond{0.88} & \marksecond{0.50} & \marksecond{0.91} & \marksecond{0.43} \\
            &  fisheye3& 0.47 & 0.70 & 0.45 & 0.78 & 0.43 & 0.85 & 0.40 \\
            &  +VC1 &0.57 & 0.72 & 0.55 & 0.78 & \marksecond{0.52} & 0.83 & \marksecond{0.47}\\
            &  +VC2 &\marksecond{0.67} & \markfirst{0.85} & \marksecond{0.58} & \markfirst{0.89} & \marksecond{0.52} & \markfirst{0.92} & 0.46 \\
            &  360 &\markfirst{0.71} & 0.82 & \markfirst{0.70} & 0.86 & \markfirst{0.67} & 0.89 & \markfirst{0.60}\\
            \hline
            \multirow{4}{*}{CosPlace~\cite{Berton_CVPR_2022_CosPlace}}
            &  pinhole &0.12 & 0.19 & 0.11 & 0.25 & 0.11 & 0.34 & 0.11\\
            &  +VC1 &0.17 & 0.27 & 0.18 & 0.32 & 0.18 & 0.39 & 0.17 \\
            &  +VC2 &\marksecond{0.29} & \marksecond{0.36} & \marksecond{0.29} & \marksecond{0.42} & \marksecond{0.28} & \marksecond{0.51} & \marksecond{0.28} \\
            &  fisheye1&0.20 & 0.32 & 0.21 & 0.40 & 0.20 & 0.49 & 0.20 \\
            &  +VC1 &0.30 & 0.39 & 0.30 & 0.43 & 0.29 & 0.49 & 0.28\\
            &  +VC2 & \marksecond{0.58} & \marksecond{0.80} & \marksecond{0.47} & \marksecond{0.87} & \marksecond{0.39} & \marksecond{0.91} & \marksecond{0.32} \\
            &  fisheye2&0.23 & 0.35 & 0.23 & 0.42 & 0.23 & 0.51 & 0.22 \\
            &  +VC1 & 0.32 & 0.40 & 0.32 & 0.45 & 0.31 & 0.51 & 0.29  \\
            &  +VC2 &\marksecond{0.60} & \marksecond{0.81} & \marksecond{0.48} & \marksecond{0.87} & \marksecond{0.41} & \marksecond{0.91} & \marksecond{0.34}\\
            &  fisheye3&0.32 & 0.45 & 0.32 & 0.53 & 0.31 & 0.61 & 0.30 \\
            &  +VC1 &0.41 & 0.50 & 0.41 & 0.55 & 0.41 & 0.60 & \marksecond{0.38}\\
            &  +VC2 &\marksecond{0.62} & \marksecond{0.82} & \marksecond{0.51} & \marksecond{0.88} & \marksecond{0.44} & \marksecond{0.91} & 0.37 \\
            &  360 &\markfirst{0.85} & \markfirst{0.89} & \markfirst{0.84} & \markfirst{0.90} & \markfirst{0.83} & \markfirst{0.93} & \markfirst{0.78} \\
            \hline
            \multirow{4}{*}{OpenIBL~\cite{ge2020self}}
            &  pinhole &0.17 & 0.33 & 0.17 & 0.43 & 0.16 & 0.55 & 0.16\\
            &  +VC1 &0.17 & 0.31 & 0.17 & 0.40 & 0.17 & 0.52 & 0.16 \\
            &  +VC2 &\marksecond{0.44} & \marksecond{0.59} & \marksecond{0.43} & \marksecond{0.66} & \marksecond{0.42} & \marksecond{0.75} & \marksecond{0.39} \\
            &  fisheye1&0.30 & 0.50 & 0.30 & 0.59 & 0.28 & 0.69 & 0.26\\
            &  +VC1 &0.47 & 0.61 & 0.45 & 0.70 & 0.44 & 0.78 & \marksecond{0.40} \\
            &  +VC2 &\marksecond{0.66} & \marksecond{0.83} & \marksecond{0.55} & \marksecond{0.88} & \marksecond{0.48} & \marksecond{0.91} & \marksecond{0.40} \\
            &  fisheye2&0.34 & 0.52 & 0.33 & 0.62 & 0.31 & 0.71 & 0.29 \\
            &  +VC1 &0.50 & 0.65 & 0.49 & 0.72 & 0.47 & 0.79 & \marksecond{0.43}\\
            &  +VC2 &\marksecond{0.67} & \marksecond{0.84} & \marksecond{0.57} & \marksecond{0.89} & \marksecond{0.50} & \markfirst{0.92} & 0.42\\
            &  fisheye3&0.46 & 0.65 & 0.44 & 0.73 & 0.42 & 0.80 & 0.39 \\
            &  +VC1 &0.61 & 0.74 & \marksecond{0.59} & 0.79 & \marksecond{0.56} & 0.84 & \marksecond{0.51} \\
            &  +VC2 &\marksecond{0.70} & \markfirst{0.86} & \marksecond{0.59} & \markfirst{0.90} & 0.52 & \markfirst{0.92} & 0.44 \\
            &  360 & \markfirst{0.80} & 0.85 & \markfirst{0.77} & 0.87 & \markfirst{0.74} & 0.89 & \markfirst{0.65}  \\
            \hline
            \multirow{4}{*}{AP-GeM~\cite{GARL17}}
            &  pinhole &0.18 & 0.34 & 0.18 & 0.42 & 0.17 & 0.52 & 0.17 \\
            &  +VC1 & 0.21 & 0.36 & 0.21 & 0.45 & 0.21 & 0.56 & 0.20 \\
            &  +VC2 & \marksecond{0.42} & \marksecond{0.59} & \marksecond{0.41} & \marksecond{0.67} & \marksecond{0.40} & \marksecond{0.76} & \marksecond{0.37} \\
            &  fisheye1&0.33 & 0.52 & 0.32 & 0.61 & 0.31 & 0.71 & 0.29 \\
            &  +VC1 &0.38 & 0.54 & 0.37 & 0.63 & 0.36 & 0.72 & 0.34 \\
            &  +VC2 & \marksecond{0.57} & \marksecond{0.80} & \marksecond{0.49} & \marksecond{0.86} & \marksecond{0.45} & \markfirst{0.90} & \marksecond{0.39} \\
            &  fisheye2&0.35 & 0.56 & 0.34 & 0.65 & 0.33 & 0.73 & 0.32 \\
            &  +VC1 &0.40 & 0.57 & 0.39 & 0.65 & 0.38 & 0.73 & 0.36  \\
            &  +VC2 & \marksecond{0.58} & \markfirst{0.81} & \marksecond{0.51} & \markfirst{0.87} & \marksecond{0.46} & \markfirst{0.90} & \marksecond{0.40} \\
            &  fisheye3& 0.42 & 0.62 & 0.39 & 0.70 & 0.38 & 0.77 & 0.36  \\
            &  +VC1 &0.47 & 0.62 & 0.45 & 0.70 & 0.43 & 0.78 & \marksecond{0.41} \\
            &  +VC2 &\marksecond{0.58} & \markfirst{0.81} & \marksecond{0.51} & \markfirst{0.87} & \marksecond{0.46} & \markfirst{0.90} & \marksecond{0.41} \\
            &  360 &\markfirst{0.71} & 0.80 & \markfirst{0.68} & 0.85 & \markfirst{0.64} & 0.88 & \markfirst{0.58} \\
            \hline
        \end{tabular}
    }
    \caption{\textit{Piatrium}. Image retrieval results based on 360$^\circ$ reference database for the top $k$ retrieved images,
$k$ = 1, 5, 10, 20. \marksecond{\footnotesize $\#$} indicates the highest value  of R@$k$ and P@$k$ for each device w and w/o virtual cameras (VC1, VC2). Best results for all devices of R@$k$ and P@$k$ are in bold with \markfirst{\footnotesize $\#$}.}
     \label{tab:ir_piatrium}
\end{table}

\begin{table*}[]
\centering
\setlength{\tabcolsep}{1pt} %
\resizebox{2\columnwidth}{!}{
\begin{tabular}{cc|ccc|ccc|ccc|ccc|c}
\hline \multicolumn{2}{c}{ Scene } & \multicolumn{13}{c}{ Day } \\
\hline \multicolumn{2}{c}{ Concourse } & \multicolumn{13}{c}{ Query camera } \\
\hline Retrieval & local Matching& pinhole & + VC1 & +VC2 & fisheye1 & + VC1 & +VC2 &  fisheye2 & + VC1 & +VC2 &fisheye3 & + VC1 & +VC2 & 360\\ \hline
 \multirow{4}{*}{ NetVLAD } & SIFT + NN &4.8/9.0/13.4&9.0/12.4/16.3&\marksecond{15.9/20.2/24.8}&2.0/4.5/10.5&4.7/\marksecond{9.1}/15.3&\marksecond{3.9}/9.0/\marksecond{17.8}&1.9/4.7/10.2&4.2/8.7/15.9&\marksecond{5.3/10.9/19.8}&3.8/8.3/15.6&6.3/11.3/18.3&\marksecond{6.7/12.6/21.7}&\markfirst{19.2/29.0/48.9} \\
 & DISK + LG & 7.7/14.5/27.6&12.6/19.2/27.5&\markfirst{16.4}/\marksecond{23.2/33.6}&1.6/5.0/21.4&3.0/8.3/25.2&\marksecond{4.4/9.8/32.8}&1.3/4.9/23.8&3.9/9.2/26.5&\marksecond{4.6/11.3/33.1}&4.8/10.5/32.0&\marksecond{4.6/13.0}/34.3&4.5/12.7/\marksecond{40.1}&\markfirst{14.3/25.8/58.9} \\
 & SP + LG &11.0/19.6/35.0&14.9/22.0/33.6&\markfirst{22.3/31.7}\marksecond{/47.4}&2.7/7.1/22.8&\marksecond{4.3}/10.5/29.6&4.2/\marksecond{12.6/38.1}&2.3/8.4/26.3&4.4/11.9/29.5&\marksecond{6.1/15.9/40.6}&3.6/9.9/30.9&\marksecond{6.2}/15.5/37.1&5.4/\marksecond{16.9/44.8}&\markfirst{17.7/31.7/56.8}\\
 & SP + SG &11.2/19.3/33.1&14.7/23.1/32.6&\markfirst{22.4/31.5}\marksecond{/46.7}&2.0/5.5/20.3&2.7/9.4/27.3&\marksecond{4.5/11.8/31.3}&2.2/6.4/21.4&\marksecond{3.9}/9.9/25.9&3.6/\marksecond{11.8/35.2}&4.2/10.5/26.5&\marksecond{6.0}/13.7/33.1&5.9/\marksecond{14.8/37.5}&14.8/26.5/\markfirst{53.0}\\
\hline
 \multirow{4}{*}{ CosPlace } & SIFT + NN & 6.2/8.8/12.4&6.0/8.9/12.0&\marksecond{10.4/13.1/15.4}&1.4/3.0/6.8&2.9/5.5/10.3&\marksecond{3.6/7.2/13.2}&1.9/4.0/8.3&3.5/7.0/11.1&\marksecond{4.0/7.8/13.7}&2.9/6.0/12.1&\marksecond{5.2}/8.9/15.3&5.1/\marksecond{9.2/17.8}&\markfirst{22.8/34.1/55.6}\\
 & DISK + LG &7.7/13.5/24.4&7.8/12.8/21.7&\marksecond{9.6/14.2/21.8}&1.2/3.4/16.0&\marksecond{2.7}/6.3/20.7&2.5/\marksecond{7.4/23.3}&1.6/4.3/18.8&2.1/7.0/19.9&\marksecond{4.0/8.0/25.4}&2.1/6.2/25.5&3.9/8.9/28.9&\marksecond{4.6/9.9/32.0}&\markfirst{14.3/27.5/59.2} \\
 & SP + LG &9.5/17.9/28.4&11.0/18.0/25.9&\marksecond{13.4/19.3/29.8}&1.5/4.8/16.3&2.0/6.9/19.0&\marksecond{3.5/9.4/28.4}&1.4/4.7/18.9&\marksecond{3.0}/7.7/20.7&\marksecond{3.0/9.4/30.7}&3.8/8.7/24.5&4.7/11.8/28.5&\marksecond{6.1/14.1/35.4}&\markfirst{18.4/33.1/58.7} \\
 & SP + SG &10.5/17.4/26.7&10.5/16.3/23.9&\marksecond{13.4/20.0/29.7}&0.8/4.5/15.1&\marksecond{1.8}/5.1/18.5&1.6/\marksecond{7.5/23.4}&1.6/4.9/14.1&2.5/7.7/18.1&\marksecond{2.6/8.7/24.6}&2.5/6.7/20.4&\marksecond{4.6/11.0}/26.9&3.8/9.9/\marksecond{29.1}&\markfirst{16.4/30.5/58.9} \\
\hline \multicolumn{2}{c}{} & \multicolumn{13}{c}{ Night } \\
\hline 
 \multirow{4}{*}{ NetVLAD } & SIFT + NN & 1.4/3.2/6.6&2.8/4.4/5.5&\marksecond{7.1/9.5/10.9}&1.2/2.2/5.1&2.5/4.9/7.3&\marksecond{3.3/6.3/9.2}&1.0/2.1/5.3&2.4/4.8/8.0&\marksecond{3.9/6.7/10.8}&1.7/4.7/7.2&\marksecond{5.5}/9.4/13.3&5.3/\marksecond{9.8/15.3}&\markfirst{21.6/29.0/36.4}\\
 & DISK + LG &5.0/11.5/19.7&6.5/10.7/16.1&\marksecond{10.4/17.2/25.7}&1.1/4.4/13.4&2.4/6.6/17.6&\marksecond{2.8/8.9/21.7}&1.3/4.6/14.5&\marksecond{2.2}/5.4/17.7&\marksecond{2.2/9.0/27.1}&2.7/7.2/22.3&3.5/10.8/27.2&\marksecond{5.4/12.5/31.3}&\markfirst{17.7/32.1/55.1}\\
 & SP + LG &6.2/14.2/27.4&8.7/14.4/21.8&\marksecond{16.0/24.3/34.4}&1.7/4.9/15.7&3.6/10.1/24.0&\marksecond{4.3/12.5/29.8}&2.4/7.1/18.7&4.6/11.8/22.1&\marksecond{4.7/14.9/33.9}&3.4/8.8/25.3&4.4/12.8/30.0&\marksecond{5.5/15.6/38.2}&\markfirst{13.2/26.8/49.6}\\
 & SP + SG & 6.3/13.9/24.2&7.8/14.6/20.5&\markfirst{14.3}/\marksecond{23.2/33.1}&0.9/3.6/10.8&\marksecond{2.5}/7.3/17.4&2.3/\marksecond{10.3/24.6}&1.6/5.2/15.1&2.5/6.9/19.6&\marksecond{3.1/10.7/25.2}&1.8/6.8/18.4&3.9/9.5/23.5&\marksecond{4.9/13.9/32.1}&14.2/\markfirst{27.8/50.8}\\
\hline
 \multirow{4}{*}{ CosPlace } & SIFT + NN & 1.7/2.9/5.8&1.7/2.9/4.7&\marksecond{2.8/3.4/4.9}&0.4/1.1/2.5&1.3/2.5/4.4&\marksecond{2.9/4.4/6.4}&0.5/1.8/3.7&1.2/2.5/4.6&\marksecond{2.5/5.6/8.2}&1.9/4.0/5.8&1.8/4.3/6.8&\marksecond{4.3/7.3/11.1}&\markfirst{23.9/31.1/39.9} \\
 & DISK + LG &4.4/8.5/15.7&4.4/8.5/14.3&\marksecond{6.1/8.3/12.9}&1.0/3.3/11.1&1.6/4.0/12.5&\marksecond{1.9/5.6/18.0}&1.2/3.7/12.9&1.2/3.8/12.1&\marksecond{1.8/6.4/19.3}&2.1/5.4/15.1&2.6/7.2/20.5&\marksecond{4.7/10.7/28.7}&\markfirst{17.5/34.0/59.9} \\
 & SP + LG & 5.2/11.8/20.2&4.8/9.8/16.4&\marksecond{7.2/10.8/19.6}&1.2/2.8/12.7&2.0/5.4/14.5&\marksecond{3.3/9.0/24.6}&1.6/3.7/15.7&1.9/4.7/13.4&\marksecond{4.0/11.6/27.1}&2.4/5.8/17.5&3.4/7.5/19.4&\marksecond{5.4/14.1/30.5}&\markfirst{14.2/29.8/55.3} \\
 & SP + SG & 5.3/11.4/19.4&5.0/9.7/16.1&\marksecond{7.0/12.0/18.6}&0.8/2.9/11.4&1.3/4.1/10.8&\marksecond{2.1/7.2/19.7}&1.4/3.5/11.7&1.3/4.5/11.2&\marksecond{2.4/7.6/21.0}&1.5/4.7/13.5&2.7/7.4/16.4&\marksecond{3.6/8.3/24.2}&\markfirst{14.0/28.2/52.7} \\
 \hline
\end{tabular}
}
\caption{\textit{Concourse}. Local matching localization results. Percentage of predictions with high (0.25m, $2^{\circ}$), medium (0.5m, $5^{\circ}$), and low (5m, $10^{\circ}$) accuracy~\cite{sattler2018benchmarking} (higher is better). \marksecond{\footnotesize  $\#$} indicates the highest value for each device w and w/o virtual cameras (VC1, VC2) of each accuracy level. The best results for all devices of each accruacy level are in bold with \markfirst{\footnotesize  $\#$}.}
\label{tab:bs_lm_concourse}
\end{table*}

\begin{table*}[]
\centering
\setlength{\tabcolsep}{1pt} %
\resizebox{2\columnwidth}{!}{
\begin{tabular}{cc|ccc|ccc|ccc|ccc|c}
\hline \multicolumn{2}{c}{ Scene } & \multicolumn{13}{c}{ Day } \\
\hline \multicolumn{2}{c}{ Hall } & \multicolumn{13}{c}{ Query camera } \\
\hline Retrieval & local Matching& pinhole & + VC1 & +VC2 & fisheye1 & + VC1 & +VC2 &  fisheye2 & + VC1 & +VC2 &fisheye3 & + VC1 & +VC2 & 360\\ \hline
 \multirow{4}{*}{ NetVLAD } & SIFT + NN &8.9/13.8/20.7&11.4/16.3/20.6&\marksecond{35.8/45.1/52.2}&5.3/11.6/25.7&11.1/21.8/37.3&\marksecond{12.6/24.3/42.7}&7.6/14.0/27.5&12.5/24.8/41.3&\marksecond{14.1/25.6/44.4}&10.6/19.6/37.5&\marksecond{16.4/32.3/50.2}&15.1/28.4/47.7&\markfirst{36.0/54.7/76.0} \\
 & DISK + LG &11.4/17.8/27.3&13.4/18.0/22.4&\marksecond{27.2/36.5/44.3}&3.8/9.9/26.7&7.3/19.4/40.4&\marksecond{8.4/21.6/43.6}&3.9/11.0/30.5&10.0/23.2/44.3&\marksecond{9.4/22.8/46.7}&8.0/19.3/41.1&\marksecond{14.7/31.8/56.0}&11.7/27.6/53.7&\markfirst{36.0/57.2/80.6} \\
 & SP + LG & 14.4/22.4/33.7&14.5/20.6/26.3&\markfirst{34.0}/\marksecond{44.7/53.6}&4.1/9.8/27.6&8.0/19.7/41.9&\marksecond{8.5/22.3/47.6}&3.9/11.8/32.0&9.7/23.9/44.9&\marksecond{9.7/24.5/50.4}&7.7/19.7/42.7&11.1/\marksecond{28.9/54.9}&\marksecond{11.4}/27.4/53.3&32.9/\markfirst{55.4/75.3 }\\
 & SP + SG &14.1/22.4/33.7&15.1/21.0/27.2&\markfirst{36.7}/\marksecond{50.4/61.4}&3.2/8.7/26.0&6.3/17.3/37.7&\marksecond{6.2/18.4/42.9}&3.7/10.4/28.5&8.3/20.7/43.5&\marksecond{7.2/20.4/44.9}&6.3/15.9/39.5&\marksecond{11.2/27.3/53.3}&10.6/24.2/47.7&27.5/\markfirst{49.8/73.6}\\
\hline
 \multirow{4}{*}{ CosPlace } & SIFT + NN & 3.9/6.3/9.8&9.0/12.8/16.2&\marksecond{15.1/20.4/25.2}&3.0/7.3/14.5&7.0/12.7/22.5&\marksecond{8.4/18.3/32.1}&3.7/7.3/16.7&8.0/14.6/23.4&\marksecond{8.2/17.4/32.8}&7.7/15.0/28.3&\marksecond{11.3}/20.7/34.8&11.0/\marksecond{22.0/39.3}&\markfirst{37.1/54.7/77.1} \\
 & DISK + LG &6.2/9.0/15.9&10.4/15.3/20.3&\marksecond{14.0/19.4/24.6}&1.7/5.0/16.6&4.9/11.4/23.7&\marksecond{6.1/14.4/33.3}&2.8/7.3/19.7&\marksecond{6.3}/14.3/28.1&5.7/\marksecond{14.9/36.3}&6.2/14.1/31.8&\marksecond{10.6/21.9}/41.3&9.6/20.4/\marksecond{44.6}&\markfirst{36.7/57.3/79.8} \\
 & SP + LG &5.8/10.4/18.8&10.4/15.5/20.7&\marksecond{16.1/22.0/27.0}&1.7/6.5/17.4&4.1/10.8/23.4&\marksecond{5.1/14.6/34.8}&2.8/7.3/19.2&5.1/13.3/27.3&\marksecond{5.5/15.5/38.0}&5.7/13.9/30.9&8.3/20.9/39.5&\marksecond{9.0/22.8/44.4}&\markfirst{31.6/52.7/77.0}\\
 & SP + SG & 6.0/10.4/19.1&11.6/16.5/22.2&\marksecond{17.9/24.2/31.1}&2.0/5.8/15.9&3.7/10.4/23.5&\marksecond{4.4/12.6/32.3}&2.3/6.8/18.8&4.4/12.4/25.4&\marksecond{5.3/13.9/34.2}&4.1/12.2/29.7&\marksecond{8.1}/18.6/38.2&7.1/\marksecond{18.7/40.4}&\markfirst{29.4/49.9/73.6}\\
\hline \multicolumn{2}{c}{} & \multicolumn{13}{c}{ Night } \\\hline
 \multirow{4}{*}{ NetVLAD } & SIFT + NN &\marksecond{0.0}/0.0/0.1&\marksecond{0.0}/0.0/0.0&\marksecond{0.0/0.4/0.7}&0.0/0.2/0.8&0.1/0.5/1.3&\marksecond{0.2/0.6/1.5}&0.0/0.1/0.8&0.0/0.4/1.2&\marksecond{0.2/0.7/1.9}&0.3/0.7/1.6&0.1/0.8/2.1&\marksecond{0.3/1.0/2.0}&\markfirst{2.1/5.3/13.5}\\
 & DISK + LG &0.4/1.6/4.5&0.5/1.3/2.5&\marksecond{1.9/4.7/8.2}&0.4/1.5/6.3&0.2/1.6/6.4&\marksecond{0.3/3.0/12.8}&0.3/1.8/8.2&0.4/2.8/9.2&\marksecond{0.7/3.9/14.8}&0.6/4.1/14.0&1.1/4.4/16.3&\marksecond{1.2/6.3/22.1}&\markfirst{7.2/22.1/53.5}\\
 & SP + LG & 0.6/2.1/6.7&0.6/1.7/3.9&\marksecond{2.1/5.4/10.5}&0.5/2.2/10.5&0.9/3.4/11.2&\marksecond{1.1/5.1/20.4}&0.4/2.6/12.1&0.9/4.1/14.0&\marksecond{1.2/5.6/23.9}&0.8/4.3/16.5&\marksecond{1.7}/6.6/21.3&\marksecond{1.7/7.9/29.2}&\markfirst{6.1/21.6/54.6} \\
 & SP + SG &0.6/2.4/6.2&0.7/1.5/4.2&\marksecond{2.5/6.9/13.1}&0.4/1.8/8.5&0.4/2.2/10.0&\marksecond{0.6/4.1/18.9}&0.4/2.2/9.8&0.5/3.0/11.9&\marksecond{1.0/5.4/22.2}&0.8/4.3/15.0&\marksecond{1.2}/5.6/18.5&1.0/\marksecond{6.2/26.4}&\markfirst{5.7/17.6/50.1} \\
\hline
 \multirow{4}{*}{ CosPlace } & SIFT + NN & 0.0/0.1/0.2&0.0/0.0/0.0&\marksecond{0.1/0.1/0.4}&0.0/0.1/0.2&0.0/0.0/0.1&\marksecond{0.1/0.3/0.9}&0.0/0.3/0.5&0.0/0.0/0.0&0.0/0.2/1.1&\marksecond{0.0}/0.2/0.8&\marksecond{0.0}/0.2/0.6&\marksecond{0.0/0.6/1.7}&\markfirst{2.1/6.7/14.9} \\
 & DISK + LG &0.8/1.8/4.1&0.6/1.4/3.7&\marksecond{1.2/3.3/6.8}&0.3/1.3/4.9&0.0/0.7/3.5&\marksecond{0.7/2.6/10.0}&0.3/1.6/6.6&0.1/0.8/4.2&\marksecond{0.3/2.3/10.5}&0.2/2.2/11.0&0.7/2.7/9.2&\marksecond{0.6/4.5/15.4}&\markfirst{7.0/23.6/62.3} \\
 & SP + LG &0.5/1.8/5.2&0.7/1.6/4.0&\marksecond{1.5/4.5/8.4}&0.3/1.6/9.0&0.3/0.9/4.5&\marksecond{0.8/3.5/16.0}&0.1/1.0/8.8&0.1/1.2/5.3&\marksecond{1.0/4.2/17.0}&0.6/2.9/13.5&0.9/3.7/12.2&\marksecond{1.4/6.1/21.4}&\markfirst{5.2/22.1/64.7} \\
 & SP + SG & 0.6/1.5/5.1&0.6/2.1/4.4&\marksecond{2.5/5.2/10.8}&0.2/1.5/6.7&0.1/1.1/3.7&\marksecond{0.7/3.6/14.5}&0.5/1.8/7.4&0.1/1.0/4.7&\marksecond{0.8/3.7/16.1}&0.4/3.0/11.6&0.8/3.5/10.8&\marksecond{1.0/4.3/19.6}&\markfirst{5.8/20.1/60.9} \\
 \hline
\end{tabular}
}
\caption{\textit{Hall}. Local matching localization results. Percentage of predictions with high (0.25m, $2^{\circ}$), medium (0.5m, $5^{\circ}$), and low (5m, $10^{\circ}$) accuracy~\cite{sattler2018benchmarking} (higher is better). \marksecond{\footnotesize  $\#$} indicates the highest value for each device w and w/o virtual cameras (VC1, VC2) of each accuracy level. The best results for all devices of each accruacy level are in bold with \markfirst{\footnotesize  $\#$}.}
\label{tab:bs_lm_hall}
\end{table*}

\begin{table*}[]
\centering
\setlength{\tabcolsep}{1pt} %
\resizebox{2\columnwidth}{!}{
\begin{tabular}{cc|ccc|ccc|ccc|ccc|c}
\hline \multicolumn{2}{c}{ Scene } & \multicolumn{13}{c}{ Day } \\
\hline \multicolumn{2}{c}{ Atrium } & \multicolumn{13}{c}{ Query camera } \\
\hline Retrieval & local Matching& pinhole & + VC1 & +VC2 & fisheye1 & + VC1 & +VC2 &  fisheye2 & + VC1 & +VC2 &fisheye3 & + VC1 & +VC2 & 360\\ \hline
 \multirow{4}{*}{ NetVLAD } & SIFT + NN & 0.9/2.1/6.9&2.2/4.6/10.1&\marksecond{6.1/10.5/20.5}&0.4/1.0/6.5&1.3/3.1/14.5&\marksecond{2.2/6.0/20.2}&1.3/2.2/8.7&2.2/5.4/16.8&\marksecond{2.4/5.4/21.6}&1.4/3.7/13.5&\marksecond{2.8/6.5}/21.5&2.5/6.3/\marksecond{26.5}&\markfirst{12.8/24.5/56.7} \\
 & DISK + LG &2.5/6.9/24.7&3.5/9.6/23.5&\marksecond{6.5/14.8/35.7}&0.2/1.3/11.1&0.8/3.1/22.4&\marksecond{1.0/4.8/30.7}&0.5/1.5/14.6&1.0/\marksecond{4.0}/27.6&\marksecond{1.1}/3.9/\marksecond{30.2}&0.9/3.0/23.5&1.3/5.0/33.5&\marksecond{2.5/7.1/38.2}& \markfirst{7.9/17.4/64.1}\\
 & SP + LG &3.5/8.9/29.3&4.7/11.4/24.2&\markfirst{10.7}/\marksecond{21.3/37.2}&0.2/1.5/14.3&1.6/5.1/28.2&\marksecond{2.0/6.9/35.1}&0.8/3.1/17.5&1.5/5.7/30.9&\marksecond{2.1/7.3/38.6}&2.2/5.6/26.2&2.6/8.3/41.0&\marksecond{3.1/9.9/46.6}& 10.5/\markfirst{24.3/65.4} \\
 & SP + SG & 4.3/9.5/29.7&5.8/12.5/26.6&\markfirst{12.7}/\marksecond{24.7/46.6}&0.7/2.1/14.1&1.0/4.6/28.1&\marksecond{1.5/6.2/34.6}&0.4/1.5/13.6&\marksecond{2.1}/5.3/30.2&1.8/\marksecond{6.7/36.9}&1.7/4.3/22.8&\marksecond{2.5}/7.8/33.9&\marksecond{2.5/8.3/40.7}&10.2/\markfirst{23.3/59.5}\\
\hline
 \multirow{4}{*}{ CosPlace } & SIFT + NN & 0.7/2.2/6.5&1.3/2.9/8.8&\marksecond{3.0/5.7/13.7}&0.3/0.8/5.4&1.2/2.9/11.4&\marksecond{1.6/3.7/15.7}&0.2/1.3/5.2&1.0/\marksecond{3.0}/11.3&\marksecond{1.2}/2.7/\marksecond{15.4}&0.9/2.3/8.8&\marksecond{2.0}/4.6/18.2&1.9/\marksecond{5.1/19.4}&\markfirst{12.8/23.9/60.7} \\
 & DISK + LG &1.8/4.9/19.0&3.2/8.1/24.5&\marksecond{4.6/10.6/28.0}&0.0/0.5/8.6&\marksecond{0.5}/1.8/17.5&0.3/\marksecond{2.6/23.0}&0.4/1.3/10.2&\marksecond{0.6}/1.9/18.1&0.5/\marksecond{2.3/23.1}&0.5/1.7/15.5&\marksecond{1.0/4.6}/26.3&\marksecond{1.0}/3.3/\marksecond{31.7}&\markfirst{9.8/19.8/71.5} \\
 & SP + LG & 2.3/7.6/23.9&3.5/9.0/23.7&\marksecond{6.7/13.1/27.3}&0.4/1.2/9.6&0.9/2.9/20.1&\marksecond{1.4/4.2/27.0}&0.2/1.7/12.3&1.0/2.9/21.8&\marksecond{1.5/4.5/29.9}&0.5/2.7/16.4&\marksecond{2.7}/6.1/32.5&1.8/\marksecond{6.9/34.9}&\markfirst{12.2/26.7/72.2} \\
 & SP + SG & 3.0/8.2/22.2&3.3/10.2/24.8&\marksecond{7.7/15.6/33.4}&0.5/1.5/9.4&1.1/2.7/18.5&\marksecond{1.3/4.0/27.7}&0.5/1.9/9.2&1.1/3.7/21.3&\marksecond{1.0/4.5/28.5}&0.7/2.1/14.4&1.7/5.8/28.1&\marksecond{2.0/6.0/33.2}&\markfirst{10.5/24.8/69.6} \\
\hline \multicolumn{2}{c}{} & \multicolumn{13}{c}{ Night } \\\hline
 \multirow{4}{*}{ NetVLAD } & SIFT + NN &0.0/0.2/0.7&0.1/0.4/0.6&\marksecond{0.9/1.6/2.5}&0.0/0.0/0.7&0.1/0.6/1.9&\marksecond{0.5/1.0/3.1}&0.2/0.4/1.1&0.2/0.8/2.5&\marksecond{0.5/1.2/3.4}&0.4/0.9/2.3&0.3/\marksecond{1.4}/3.1&\marksecond{0.5}/1.3/\marksecond{4.3}&\markfirst{3.9/6.5/15.5}\\
 & DISK + LG &1.1/4.0/13.5&1.6/3.6/10.6&\marksecond{3.9/8.3/17.0}&0.4/1.2/8.4&0.5/2.2/11.2&\marksecond{1.0/3.2/18.1}&0.2/1.4/9.7&0.6/2.7/13.3&\marksecond{1.1/4.8/21.7}&0.8/2.6/14.2&1.4/4.6/20.2&\marksecond{1.6/5.7/25.8}&\markfirst{7.1/19.4/56.4} \\
 & SP + LG &1.5/4.2/14.5&1.9/3.9/11.8&\marksecond{5.5/10.3/18.1}&0.4/1.6/11.1&0.7/3.2/17.6&\marksecond{1.6/5.4/26.5}&0.4/1.8/11.9&1.1/4.0/18.5&\marksecond{1.6/6.1/29.3}&0.7/3.8/16.0&1.6/5.5/25.0&\marksecond{1.9/7.6/32.2}&\markfirst{7.0/19.7/51.7} \\
 & SP + SG & 1.6/4.1/14.4&2.1/4.9/12.3&\marksecond{5.9/11.7/21.7}&0.3/1.6/9.8&0.7/2.4/13.7&\marksecond{1.4/5.5/24.8}&0.6/2.1/11.3&0.9/3.2/16.8&\marksecond{1.6/5.7/26.3}&0.5/2.7/15.3&1.6/4.8/21.8&\marksecond{2.1/6.8/30.4}&\markfirst{6.3/18.7/50.5}\\
\hline
 \multirow{4}{*}{ CosPlace } & SIFT + NN &0.0/0.1/0.2&0.2/0.3/0.7&\marksecond{0.6/0.9/1.4}&0.0/0.1/0.5&0.0/0.0/0.8&\marksecond{0.3/0.5/2.0}&0.0/0.1/0.7&0.1/0.3/1.0&\marksecond{0.1/0.7/2.5}&0.1/0.3/0.8&0.2/0.7/2.2&\marksecond{0.5/1.1/3.0}&\markfirst{3.5/7.9/19.0} \\
 & DISK + LG & 0.9/3.0/10.6&1.6/4.3/11.8&\marksecond{2.7/5.9/13.0}&0.2/0.8/4.9&0.3/1.3/10.7&\marksecond{0.7/2.5/15.0}&0.2/0.9/5.8&0.4/1.6/10.7&\marksecond{0.8/2.7/17.9}&0.2/1.4/9.9&0.9/3.0/16.7&\marksecond{1.4/4.8/24.6}&\markfirst{7.8/22.5/62.3} \\
 & SP + LG & 1.4/3.8/13.1&2.3/5.0/13.5&\marksecond{3.0/6.5/14.2}&0.3/1.1/7.7&0.3/1.6/12.9&\marksecond{0.8/4.0/21.7}&0.2/0.9/8.7&0.9/2.8/14.9&\marksecond{1.2/4.3/24.7}&0.6/2.3/11.6&0.9/3.5/19.2&\marksecond{1.2/5.6/29.2}&\markfirst{6.4/18.0/55.7} \\
 & SP + SG &1.5/4.1/12.4&2.4/5.5/13.5&\marksecond{3.8/8.2/16.9}&0.1/0.8/6.4&0.4/1.8/12.0&\marksecond{0.6/3.9/20.8}&0.3/1.0/6.8&0.7/2.3/11.6&\marksecond{1.0/3.9/22.2}&0.3/1.7/9.7&0.9/2.9/15.8&\marksecond{1.4/6.0/26.9}&\markfirst{6.4/17.3/53.1} \\
 \hline
\end{tabular}
}
\caption{\textit{Atrium}. Local matching localization results. Percentage of predictions with high (0.25m, $2^{\circ}$), medium (0.5m, $5^{\circ}$), and low (5m, $10^{\circ}$) accuracy~\cite{sattler2018benchmarking} (higher is better). \marksecond{\footnotesize  $\#$} indicates the highest value for each device w and w/o virtual cameras (VC1, VC2) of each accuracy level. The best results for all devices of each accruacy level are in bold with \markfirst{\footnotesize  $\#$}.}
\label{tab:bs_lm_atrium}
\end{table*}

\begin{table*}[]
\centering
\setlength{\tabcolsep}{1pt} %
\resizebox{2\columnwidth}{!}{
\begin{tabular}{cc|ccc|ccc|ccc|ccc|c}
\hline \multicolumn{2}{c}{ Scene } & \multicolumn{13}{c}{ Day } \\
\hline \multicolumn{2}{c}{ Piatrium } & \multicolumn{13}{c}{ Query camera } \\
\hline Retrieval & local Matching& pinhole & + VC1 & +VC2 & fisheye1 & + VC1 & +VC2 &  fisheye2 & + VC1 & +VC2 &fisheye3 & + VC1 & +VC2 & 360\\ \hline
 \multirow{4}{*}{ NetVLAD } & SIFT + NN & 1.2/2.5/7.9&4.1/7.6/13.1&\marksecond{10.1/16.2/27.5}&0.9/2.1/8.4&1.9/5.5/18.0&\marksecond{3.0/6.9/21.0}&1.2/2.3/8.1&3.2/7.0/18.9&\marksecond{2.2/6.0/21.6}&2.5/5.7/15.3&\marksecond{4.1/9.0/24.7}&3.7/8.0/23.2&\markfirst{13.4/24.8/51.3} \\
 & DISK + LG &2.3/6.1/18.8&4.4/9.4/20.5&\marksecond{6.8/14.2/28.5}&0.7/1.4/11.6&2.1/6.0/22.2&\marksecond{1.9/5.6/25.1}&0.8/2.3/14.5&\marksecond{2.2/6.6}/25.3&2.2/6.1/\marksecond{27.7}&1.3/5.1/22.4&\marksecond{3.1/8.9}/34.0&2.3/8.2/\marksecond{35.1}&\markfirst{10.0/22.9/60.9} \\
 & SP + LG &3.2/8.7/25.7&7.5/14.1/26.1&\markfirst{12.3}/\marksecond{21.0/33.4}&0.8/3.1/15.6&2.5/7.2/29.0&\marksecond{2.5/7.8/31.8}&0.7/3.6/17.0&3.3/8.3/31.0&\marksecond{2.6/8.4/34.7}&2.4/6.7/26.6&\marksecond{4.3/12.3/40.9}&3.7/11.7/40.5&11.9/\markfirst{26.9/59.2} \\
 & SP + SG &3.9/9.6/26.2&8.1/14.7/27.7&\markfirst{14.4/26.4}/\marksecond{43.9}&0.4/2.5/13.2&2.0/6.8/25.4&\marksecond{2.2/7.6/29.1}&0.4/2.4/14.5&2.1/6.6/26.5&\marksecond{2.3/7.0/30.0}&1.6/5.5/24.8&\marksecond{3.6/10.1}/36.1&2.7/9.6/\marksecond{36.2}&10.7/25.1/\markfirst{55.5} \\\hline
 \multirow{4}{*}{ CosPlace } & SIFT + NN & 0.7/1.1/3.8&2.6/4.5/10.0&\marksecond{4.2/7.9/15.0}&0.5/1.4/4.6&1.3/3.0/11.9&\marksecond{2.1/4.3/14.6}&0.6/1.5/5.6&1.5/3.7/12.8&\marksecond{2.1/4.7/15.6}&1.3/3.2/9.8&\marksecond{2.8/5.9}/16.3&2.3/5.3/\marksecond{19.0}&\markfirst{10.3/21.3/50.9} \\
 & DISK + LG &1.2/3.8/12.7&3.1/6.9/17.7&\marksecond{3.8/8.2/19.7}&0.4/1.2/6.1&0.9/2.7/15.6&\marksecond{1.0/3.4/21.6}&0.5/1.2/8.3&0.9/2.9/16.5&\marksecond{1.2/4.1/23.5}&1.0/3.2/14.7&\marksecond{2.0}/5.3/24.0&\marksecond{2.0/5.7/30.2}& \markfirst{9.7/22.8/61.8} \\
 & SP + LG &1.6/5.1/17.3&5.2/10.2/19.7&\marksecond{6.4/11.3/22.5}&0.3/1.5/8.8&1.6/3.9/16.9&\marksecond{1.0/4.4/26.0}&0.6/1.7/10.5&1.0/4.2/18.8&\marksecond{1.6/6.4/29.3}&1.1/3.4/17.4&\marksecond{2.7/7.5}/27.7&1.8/6.8/\marksecond{32.3}& \markfirst{12.4/27.0/64.5} \\
 & SP + SG & 2.2/5.5/16.5&5.2/10.9/20.3&\marksecond{7.4/14.2/27.8}&0.5/1.6/8.2&1.1/3.7/15.9&\marksecond{1.0/4.3/23.4}&0.6/1.9/9.4&0.9/3.4/16.1&\marksecond{1.2/4.7/24.4}&0.9/2.7/15.5&\marksecond{2.8}/6.4/25.7&2.5/\marksecond{7.3/27.5}& \markfirst{10.1/25.4/60.8} \\
\hline \multicolumn{2}{c}{ } & \multicolumn{13}{c}{ Night } \\ \hline
 \multirow{4}{*}{ NetVLAD } & SIFT + NN &\marksecond{0.1/0.1/0.1}&0.0/0.0/0.1&0.0/\marksecond{0.1/0.1}&0.0/0.1/0.3&0.0/0.1/0.4&\marksecond{0.1/0.4/0.9}&0.0/0.0/0.0&\marksecond{0.0/0.1/0.4}&\marksecond{0.0/0.1/0.4}&0.0/0.0/0.3&0.0/0.1/0.4&\marksecond{0.1/0.2/1.1}&\markfirst{0.4/1.0/2.9}\\
 & DISK + LG &0.2/0.6/3.6&\marksecond{0.4}/0.7/2.4&0.1/\marksecond{0.9/3.6}&0.0/0.1/1.6&0.0/0.4/3.1&\marksecond{0.1/0.8/5.7}&0.0/0.1/2.2&0.1/1.1/4.5&\marksecond{0.2/1.1/5.7}&0.0/0.6/4.6&0.1/1.2/7.0&\marksecond{0.2/1.6/10.9}&\markfirst{1.9/6.9/24.8}\\
 & SP + LG &0.6/1.4/5.5&0.4/1.1/2.9&\marksecond{0.8/1.6/4.4}&0.2/0.4/4.7&0.3/0.8/7.0&\marksecond{0.5/2.5/10.5}&0.1/0.4/4.4&\marksecond{0.4}/1.9/8.5&\marksecond{0.4/2.0/11.9}&0.2/1.4/7.8&0.5/2.5/11.3&\marksecond{0.9/3.3/16.6}&\markfirst{1.6/6.6/25.3}\\
 & SP + SG & 0.6/1.9/4.3&0.5/1.4/2.5&\marksecond{1.0/2.4/5.9}&0.3/0.4/3.8&0.1/0.5/5.9&\marksecond{0.2/1.5/9.1}&0.1/0.4/3.4&0.3/1.4/6.5&\marksecond{0.4/1.7/11.3}&0.2/1.2/6.6&0.5/2.2/9.4&\marksecond{0.4/2.4/14.8}&\markfirst{2.0/6.9/19.9} \\
\hline
 \multirow{4}{*}{ CosPlace } & SIFT + NN & \marksecond{0.1/0.1/0.2}&0.0/0.0/0.1&\marksecond{0.1/0.1/0.2}&\marksecond{0.0/0.0/0.3}&\marksecond{0.0/0.0/0.3}&\marksecond{0.0/0.0/0.3}&0.0/0.0/0.1&0.1/0.1/0.3&\marksecond{0.0/0.0/0.6}&0.0/\marksecond{0.2}/0.3&0.0/0.0/0.1&\marksecond{0.1}/0.1/\marksecond{0.9}&\markfirst{0.4/2.0/6.2}\\
 & DISK + LG &0.1/\marksecond{0.8/3.9}&0.1/0.3/3.0&\marksecond{0.2/0.8/3.9}&0.0/0.1/2.5&\marksecond{0.1/0.4}/2.7&\marksecond{0.1/0.4/5.7}&0.0/0.1/2.4&0.1/\marksecond{0.5}/2.8&\marksecond{0.1}/0.3/\marksecond{5.8}&0.0/0.4/4.3&\marksecond{0.1}/0.4/4.6&\marksecond{0.1/0.8/9.3}&\markfirst{2.6/8.0/39.3} \\
 & SP + LG &\marksecond{0.4/1.4/5.8}&0.3/1.5/4.4&\marksecond{0.4}/1.0/3.6&0.1/0.3/3.5&0.1/0.9/4.0&\marksecond{0.4/1.9/9.7}&0.1/0.5/3.3&\marksecond{0.3}/0.8/3.9&0.2/\marksecond{1.3/11.7}&0.1/0.6/7.1&0.4/1.7/7.0&\marksecond{0.7/2.8/14.3}&\markfirst{3.3/10.2/38.0}\\
 & SP + SG &\marksecond{0.9/1.7/4.9}&0.5/1.3/4.2&0.6/1.6/4.8&0.1/0.3/3.6&0.1/0.8/2.9&\marksecond{0.4/1.4/9.8}&0.1/0.3/2.9&0.0/0.5/4.4&\marksecond{0.3/1.8/11.3}&0.2/0.9/5.5&0.3/1.4/6.2&\marksecond{0.4/1.9/12.1}&\markfirst{2.3/9.2/35.0} \\
 \hline
\end{tabular}
}
\caption{\textit{Piatrium}. local matching localization results. Percentage of predictions with high (0.25m, $2^{\circ}$), medium (0.5m, $5^{\circ}$), and low (5m, $10^{\circ}$) accuracy~\cite{sattler2018benchmarking} (higher is better). \marksecond{\footnotesize  $\#$} indicates the highest value for each device w and w/o virtual cameras (VC1, VC2) of each accuracy level. The best results for all devices of each accruacy level are in bold with \markfirst{\footnotesize  $\#$}.}
\label{tab:bs_lm_piatrium}
\end{table*}

\subsection{Absolute Pose Regression}
The average local median translation and rotation errors over 4 scenes are shown in Figure~5 of the main paper.  We provide the results of each scene in this supplementary material in Table~\ref{tab:res_apr_concourse}, Table~\ref{tab:res_apr_hall}, Table~\ref{tab:res_apr_atrium} and Table~\ref{tab:res_apr_piatrium}. The trend is similar to the Figure~5 shown in the main paper. When we introduce images
from virtual cameras for data augmentation, PN$^{vc2}$ and MS-T$^{vc2}$
 exhibit significantly reduced translation and rotation errors across all cameras in most cases, particularly during daytime.

\begin{table*}[]
    \centering
    \setlength{\tabcolsep}{2pt} %
\resizebox{2\columnwidth}{!}{
    \begin{tabular}{lcccccccccc}
    \hline
        scene&\multicolumn{10}{c}{Concourse}\\\hline
        query & \multicolumn{2}{c}{pinhole} & \multicolumn{2}{c}{fisheye1} & \multicolumn{2}{c}{fisheye2}& \multicolumn{2}{c}{fisheye2} & \multicolumn{2}{c}{360} \\
        \cmidrule(lr){2-3} \cmidrule(lr){4-5} \cmidrule(lr){6-7}  \cmidrule(lr){8-9} \cmidrule(lr){10-11} 
        APR& PN & PN$^{vc2}$& PN & PN$^{vc2}$& PN & PN$^{vc2}$& PN & PN$^{vc2}$& PN & PN$^{vc2}$\\
        Day& 18.6/95.0 & \marksecond{7.7/26.6} & 10.7/88.7 &\marksecond{3.5/14.0}& 10.1/87.7& \marksecond{3.3}/\markfirst{13.8} &7.7/85.9& \marksecond{3.0/15.2} &2.6/39.0 &\markfirst{2.1}/\marksecond{38.5}\\
         Night& 19.8/96.2& \marksecond{11.6/46.5} & 13.3/85.5 & \marksecond{8.4/22.0}& 13.0/88.4& \marksecond{8.3}/\markfirst{21.3} &11.1/87.5 &\marksecond{8.9/20.4} &\markfirst{5.5}/\marksecond{45.6} &8.1/46.6\\
        \cmidrule(lr){2-3} \cmidrule(lr){4-5} \cmidrule(lr){6-7}  \cmidrule(lr){8-9} \cmidrule(lr){10-11} 
        APR& MS-T & MS-T$^{vc2}$& MS-T & MS-T$^{vc2}$& MS-T & MS-T$^{vc2}$& MS-T & MS-T$^{vc2}$& MS-T & MS-T$^{vc2}$\\\cmidrule(lr){2-3} \cmidrule(lr){4-5} \cmidrule(lr){6-7}  \cmidrule(lr){8-9} \cmidrule(lr){10-11} 
        Day& 17.0/69.1 & \marksecond{4.6/20.4} & 11/60.8 & \marksecond{2.0/10.5}& 10.5/59.6& \marksecond{1.9/10.3}& 9.9/63.4 &\markfirst{1.7}/\marksecond{10.6} &4.9/34.6& \marksecond{1.9}/\markfirst{10.2}\\
        Night& 22.1/74.7 & \marksecond{8.9/36.6} & 17.7/69.5& \marksecond{5.9/18.6}&17.0/68.3 & \markfirst{5.7}/\markfirst{18.1} &17.2/72.7 & \marksecond{5.8/19.2} &8.4/57.3 &\marksecond{7.0}/\markfirst{18.1}\\\hline
    \end{tabular}
    }
    \caption{\textit{Concourse}. The median translation and rotation errors (m/$^\circ$) of different APRs during daytime (lower is better). \marksecond{\footnotesize $\#$} indicates the lowest value of error for each device of APR and APR$^{vc2}$. Best results for all devices of APR and APR $^{vc2}$ are in bold with \markfirst{\footnotesize $\#$}.}
    \label{tab:res_apr_concourse}
\end{table*}

\begin{table*}[]
    \centering
    \setlength{\tabcolsep}{2pt} %
\resizebox{2\columnwidth}{!}{
    \begin{tabular}{lcccccccccc}
    \hline
        scene&\multicolumn{10}{c}{Hall}\\\hline
        query & \multicolumn{2}{c}{pinhole} & \multicolumn{2}{c}{fisheye1} & \multicolumn{2}{c}{fisheye2}& \multicolumn{2}{c}{fisheye2} & \multicolumn{2}{c}{360} \\
        \cmidrule(lr){2-3} \cmidrule(lr){4-5} \cmidrule(lr){6-7}  \cmidrule(lr){8-9} \cmidrule(lr){10-11} 
        APR& PN & PN$^{vc2}$& PN & PN$^{vc2}$& PN & PN$^{vc2}$& PN & PN$^{vc2}$& PN & PN$^{vc2}$\\
        Day& 18.6/99.4 & \marksecond{8.5/35.5}& 11.1/95.7& \marksecond{3.2}/\markfirst{19.7} &10.4/95.9& \marksecond{3.0/20.0}& 7.5/92.3& \marksecond{2.8/20.3}& 2.7/97.3 &\markfirst{1.9}/\marksecond{76.5}\\
        Night& 22.8/95.6& \marksecond{20.5/91.3}& 18.3/93.8& \marksecond{15.1/65.5} &17.9/93.6 &\marksecond{14.8}/\markfirst{63.2} &16.8/91.3& \markfirst{14.4}/\marksecond{63.8} &\marksecond{10.9}/100.1 &15.2/\marksecond{77.1}\\
        \cmidrule(lr){2-3} \cmidrule(lr){4-5} \cmidrule(lr){6-7}  \cmidrule(lr){8-9} \cmidrule(lr){10-11} 
        APR& MS-T & MS-T$^{vc2}$& MS-T & MS-T$^{vc2}$& MS-T & MS-T$^{vc2}$& MS-T & MS-T$^{vc2}$& MS-T & MS-T$^{vc2}$\\\cmidrule(lr){2-3} \cmidrule(lr){4-5} \cmidrule(lr){6-7}  \cmidrule(lr){8-9} \cmidrule(lr){10-11} 
        Day& 21.8/96.0 &\marksecond{5.0/37.9}& 14.1/94.0& \marksecond{2.0/22.7}& 13.5/92.7 &\marksecond{1.8/21.2} & 11.2/91.7& \markfirst{1.6}/\marksecond{21.9}& 11.4/88.5 &\marksecond{1.7}/\markfirst{19.6}\\
        Night&  32.1/96.1 &\marksecond{29.4/88.6}& 29.9/94.2 &\marksecond{18.6}/\markfirst{28.9} &29.7/93.6& \marksecond{18.4/76.5}&  31.8/93.1& \marksecond{18.5/76.0}& 22.3/94.4 &\markfirst{14.7}/\marksecond{75.0}
        \\\hline
    \end{tabular}
    }
    \caption{\textit{Hall}. The median translation and rotation errors (m/$^\circ$) of different APRs during daytime (lower is better). \marksecond{\footnotesize $\#$} indicates the lowest value of error for each device of APR and APR$^{vc2}$. Best results for all devices of APR and APR $^{vc2}$ are in bold with \markfirst{\footnotesize $\#$}.}
    \label{tab:res_apr_hall}
\end{table*}

\begin{table*}[]
    \centering
    \setlength{\tabcolsep}{2pt} %
\resizebox{2\columnwidth}{!}{
    \begin{tabular}{lcccccccccc}
    \hline
        scene&\multicolumn{10}{c}{Atrium}\\\hline
        query & \multicolumn{2}{c}{pinhole} & \multicolumn{2}{c}{fisheye1} & \multicolumn{2}{c}{fisheye2}& \multicolumn{2}{c}{fisheye2} & \multicolumn{2}{c}{360} \\
        \cmidrule(lr){2-3} \cmidrule(lr){4-5} \cmidrule(lr){6-7}  \cmidrule(lr){8-9} \cmidrule(lr){10-11} 
        APR& PN & PN$^{vc2}$& PN & PN$^{vc2}$& PN & PN$^{vc2}$& PN & PN$^{vc2}$& PN & PN$^{vc2}$\\
        Day& 15.5/85.5& \marksecond{10.8/37.0}& 11.8/79.9& \marksecond{5.0}/\markfirst{17.3} &11.3/80.4& \marksecond{4.7}/\markfirst{17.3} & 10.1/79.8 &\marksecond{4.2/17.7} &4.5/82.6& \markfirst{3.4}/\marksecond{78.4}\\
        Night& 19.8/90.8 & \marksecond{16.1/83.1} & 16.4/87.3 & \marksecond{10.0/46.5} &16.1/86.7& \marksecond{ 9.9/46.6}& 15.7/83.5 & \marksecond{10.3/47.1}&  \markfirst{8.7/22.8}&  10.8/30.7\\
        \cmidrule(lr){2-3} \cmidrule(lr){4-5} \cmidrule(lr){6-7}  \cmidrule(lr){8-9} \cmidrule(lr){10-11} 
        APR& MS-T & MS-T$^{vc2}$& MS-T & MS-T$^{vc2}$& MS-T & MS-T$^{vc2}$& MS-T & MS-T$^{vc2}$& MS-T & MS-T$^{vc2}$\\\cmidrule(lr){2-3} \cmidrule(lr){4-5} \cmidrule(lr){6-7}  \cmidrule(lr){8-9} \cmidrule(lr){10-11} 
        Day& 20.1/77.7& \marksecond{10.5/46.9}& 14.5/73.7& \marksecond{5.4/28.0}& 14.0/71.7& \marksecond{5.1}/\markfirst{27.2}& 12.5/65.6& \markfirst{4.6}/\marksecond{29.1}& 13.2/54.7& \marksecond{5.2/41.1}\\
        Night& 24.8/84.7& \marksecond{17.7/79.7}& 17.8/80.5& \marksecond{10.9/72.0} &17.0/80.1& \marksecond{10.4/72.3}& 15.1/80.0& \marksecond{10.3/73.5}& \markfirst{5.0/46.3} &8.5/68.6\\\hline
    \end{tabular}
    }
    \caption{\textit{Atrium}. The median translation and rotation errors (m/$^\circ$) of different APRs during daytime (lower is better). \marksecond{\footnotesize $\#$} indicates the lowest value of error for each device of APR and APR$^{vc2}$. Best results for all devices of APR and APR $^{vc2}$ are in bold with \markfirst{\footnotesize $\#$}.}
    \label{tab:res_apr_atrium}
\end{table*}

\begin{table*}[]
    \centering
    \setlength{\tabcolsep}{2pt} %
\resizebox{2\columnwidth}{!}{
    \begin{tabular}{lcccccccccc}
    \hline
        scene&\multicolumn{10}{c}{Piatrium}\\\hline
        query & \multicolumn{2}{c}{pinhole} & \multicolumn{2}{c}{fisheye1} & \multicolumn{2}{c}{fisheye2}& \multicolumn{2}{c}{fisheye2} & \multicolumn{2}{c}{360} \\
        \cmidrule(lr){2-3} \cmidrule(lr){4-5} \cmidrule(lr){6-7}  \cmidrule(lr){8-9} \cmidrule(lr){10-11} 
        APR& PN & PN$^{vc2}$& PN & PN$^{vc2}$& PN & PN$^{vc2}$& PN & PN$^{vc2}$& PN & PN$^{vc2}$\\
        Day& 21.3/88.0& \marksecond{12.2/38.4} &13.8/86.5 &\marksecond{4.9}/\markfirst{18.7} &13.0/87.1& \marksecond{4.6/18.9}& 10.9/84.4& \marksecond{4.4/19.6}& 5.7/69.8& \markfirst{3.8}/\marksecond{59.5}\\
        Night& 33.2/89.7 &\marksecond{28.8/90.4}& 29.3/89.6& \markfirst{23.2}/ \marksecond{64.6} &28.8/90.4& \marksecond{23.3}/ \markfirst{64.1}& 29.2/89.8 &\marksecond{23.5/67.0} &30.6/98.5 &\marksecond{23.5/80.8} \\
        \cmidrule(lr){2-3} \cmidrule(lr){4-5} \cmidrule(lr){6-7}  \cmidrule(lr){8-9} \cmidrule(lr){10-11} 
        APR& MS-T & MS-T$^{vc2}$& MS-T & MS-T$^{vc2}$& MS-T & MS-T$^{vc2}$& MS-T & MS-T$^{vc2}$& MS-T & MS-T$^{vc2}$\\\cmidrule(lr){2-3} \cmidrule(lr){4-5} \cmidrule(lr){6-7}  \cmidrule(lr){8-9} \cmidrule(lr){10-11} 
        Day&  30.0/78.5& \marksecond{10.8/35.8}& 23.4/71.3& \marksecond{4.0/21.1} &22.7/72.7 & \marksecond{3.8}/\marksecond{20.7} &19.3/72.6& \markfirst{3.7}/ \marksecond{20.4} &11.3/52.0 & \marksecond{4.1}/\markfirst{17.6}\\
        Night&  37.7/91.3 &\marksecond{30.0/87.4} &37.3/89.1& \marksecond{24.3/75.1} &36.2/89.0 &\marksecond{23.8/74.5}& 35.7/85.6 &\marksecond{23.9/75.1} &37.2/83.0 & \markfirst{22.1}/\markfirst{53.7}\\\hline
    \end{tabular}
    }
    \caption{\textit{Piatrium}. The median translation and rotation errors (m/$^\circ$) of different APRs during daytime (lower is better). \marksecond{\footnotesize $\#$} indicates the lowest value of error for each device of APR and APR$^{vc2}$. Best results for all devices of APR and APR $^{vc2}$ are in bold with \markfirst{\footnotesize $\#$}.}
    \label{tab:res_apr_piatrium}
\end{table*}

\end{document}